\documentclass[10pt,twocolumn,letterpaper]{article}

\usepackage[pagenumbers]{iccv} %

\usepackage{arydshln}
\usepackage{flushend}
\usepackage{graphbox}
\usepackage{makecell}
\usepackage{multirow}
\usepackage{siunitx}
\usepackage{xfrac}
\usepackage{xspace}

\usepackage{etoolbox}
\renewcommand{\bfseries}{\fontseries{b}\selectfont} 
\robustify\bfseries        
\newrobustcmd{\B}{\bfseries}   

\newcommand{\dashrule}[1]{
    \addlinespace[2pt]
    \cdashline{#1}[1pt/2pt]
    \addlinespace[3pt]
}

\definecolor{celestialblue}{rgb}{0.29, 0.59, 0.82}
\newif\ifshowcomments

\showcommentstrue %
\ifshowcomments
    \usepackage[textsize=scriptsize]{todonotes}
    \newcommand{\matt}[1]{\textcolor{BrickRed}{\textbf MW: #1}\xspace}
    \newcommand{\gs}[1]{\textcolor{blue!30!ForestGreen}{\textbf{gs}: #1}\xspace}
\else
    \usepackage[disable, textsize=scriptsize]{todonotes}
    \newcommand{\matt}[1]{}
    \newcommand{\gs}[1]{}
\fi

\newcommand{\alg}{\textsc{SplArt}\xspace}

\makeatletter
\renewcommand{\paragraph}{%
  \@startsection{paragraph}{4}%
  {\z@}{0.50ex \@plus 1ex \@minus .2ex}{-1em}%
  {\normalfont\normalsize\bfseries}%
}
\makeatother

\newcommand{\nsd}[2]{$#1 \scalebox{0.75}{$\pm #2$}$}

\newif\ifforarxiv

\definecolor{iccvblue}{rgb}{0.21,0.49,0.74}
\usepackage[pagebackref,breaklinks,colorlinks,allcolors=iccvblue]{hyperref}

\def\paperID{9035} %
\def\confName{ICCV}
\def\confYear{2025}

\title{\alg: Articulation Estimation and Part-Level Reconstruction\\
with 3D Gaussian Splatting}

\author{Shengjie Lin$^\dagger$, Jiading Fang$^\dagger$, Muhammad Zubair Irshad$^\ddagger$, Vitor Campagnolo Guizilini$^\ddagger$,\\
Rares Andrei Ambrus$^\ddagger$, Greg Shakhnarovich$^\dagger{}^\ddagger$, Matthew R.\ Walter$^\dagger$\\
$^\dagger$~Toyota Technological Institute at Chicago, $\ddagger$~Toyota Research Institute\\
{\tt\small \{slin, fjd, greg, mwalter\}@ttic.edu, \{zubair.irshad, vitor.guizilini, rares.ambrus\}@tri.global}}

\begin{document}

\maketitle

\begin{abstract}
Reconstructing articulated objects prevalent in daily environments is crucial for applications in augmented/virtual reality and robotics. However, existing methods face scalability limitations (requiring 3D supervision or costly annotations), robustness issues (being susceptible to local optima), and rendering shortcomings (lacking speed or photorealism). We introduce \alg, a self-supervised, category-agnostic framework that leverages 3D Gaussian Splatting (3DGS) to reconstruct articulated objects and infer kinematics from two sets of posed RGB images captured at different articulation states, enabling real-time photorealistic rendering for novel viewpoints and articulations. \alg augments 3DGS with a differentiable mobility parameter per Gaussian, achieving refined part segmentation. A multi-stage optimization strategy is employed to progressively handle reconstruction, part segmentation, and articulation estimation, significantly enhancing robustness and accuracy. \alg exploits geometric self-supervision, effectively addressing challenging scenarios without requiring 3D annotations or category-specific priors. Evaluations on established and newly proposed benchmarks, along with applications to real-world scenarios using a handheld RGB camera, demonstrate \alg's state-of-the-art performance and real-world practicality. Code is publicly available at \url{https://github.com/ripl/splart}.
\end{abstract}

\begin{figure}[htb]
    \centering
    \includegraphics[width=\linewidth]{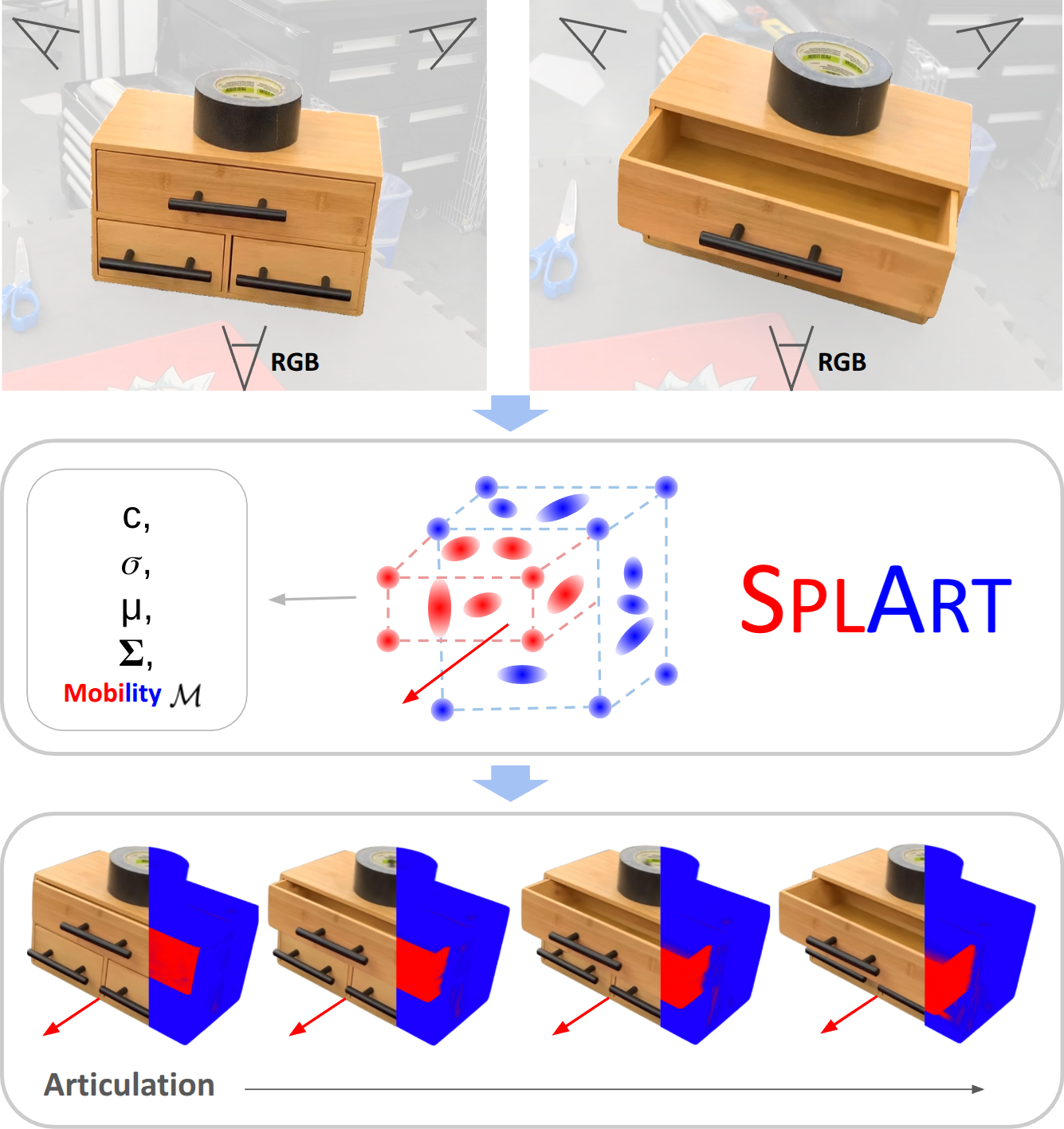}
    \caption{Given (top) RGB images of an object at two articulation states, (middle) \alg uses 3DGS to simultaneously reconstruct its static and dynamic parts and estimate the kinematic articulation model. \alg is then able to (bottom) render high-fidelity 3D reconstructions of the object along with part-level segmentations for novel articulation states, allowing for novel view synthesis.}
    \label{fig:teaser}
\end{figure}

\section{Introduction}\label{sec:intro}
Articulated objects, such as drawers, doors, and scissors, are ubiquitous in our daily lives, yet their dynamic nature poses significant challenges for 3D reconstruction---a critical task for applications in augmented/virtual reality~\cite{marchand2015pose, suzuki2022augmented}, robotics~\cite{sturm2011probabilistic, katz2013interactive, daniele17a, abbatematteo2019learning, huang2012occlusion, wang2024systematic}, and computer vision~\cite{pillai14, jain2021screwnet}. Existing methods for reconstructing articulated objects are hindered by several key limitations: they often require labor-intensive supervision (e.g., part-level segmentation or articulation annotations)~\cite{li2020category, mu2021sdf, tseng2022cla, wang2019shape2motion, jiang2022ditto}, they depend on 3D supervision that restricts practical use~\cite{li2020category, mu2021sdf, jiang2022ditto, liu2025building, weng2024neural}, they produce category-specific models that limit scalability~\cite{li2020category, mu2021sdf, tseng2022cla, wei2022self}, or they are not capable of real-time, photorealistic rendering~\cite{li2020category, mu2021sdf, tseng2022cla, wang2019shape2motion, jiang2022ditto, weng2024neural, liu2023paris, heppert2023carto, wei2022self, deng2024articulate}. To address these challenges, we introduce \alg, a novel self-supervised and category-agnostic framework that leverages 3D Gaussian Splatting (3DGS)~\cite{kerbl20233dgs} to reconstruct articulated objects from minimal input---two sets of posed RGB images at distinct articulation states. \alg reconstructs object parts and infers kinematics, enabling real-time, photorealistic rendering for novel views and articulation states.

Central to \alg is the augmentation of 3DGS~\cite{kerbl20233dgs} to include a differentiable mobility parameter for each Gaussian, which enables a more refined segmentation of static and mobile parts through gradient-based optimization. This results in enhanced reconstruction quality, while  preserving the real-time, photorealistic rendering capabilities of 3DGS---offering a speedup of more than $100\times$ over methods~\cite{liu2023paris, deng2024articulate} based on neural radiance fields~\cite{mildenhall2021nerf}.

To enhance robustness, \alg employs a multi-stage optimization strategy that decouples the part-level reconstruction and articulation estimation processes. Unlike end-to-end approaches prone to local optima~\cite{liu2023paris}, \alg first independently reconstructs each articulation state, then estimates each Gaussian's mobility parameter for part segmentation, and finally refines both the articulation and mobility estimates jointly. This structured approach ensures stable and accurate convergence, avoiding the stringent initialization requirements of existing methods, thereby providing a practical solution for challenging articulated structures.

Building on this foundation, \alg leverages geometric self-supervision to eliminate the need for manual annotations or 3D supervision. By enforcing geometric consistency between reconstructions through complementary loss formulations, \alg robustly estimates articulation parameters across diverse scenarios. This self-supervised strategy enhances scalability, enabling \alg to reconstruct a wide range of articulated objects without relying on prior structural or categorical knowledge.

Extensive evaluations on both established and newly introduced benchmarks demonstrate \alg's superior articulation accuracy and reconstruction quality, surpassing state-of-the-art methods without requiring 3D supervision. Real-world experiments further validate its practicality, showcasing successful reconstructions of diverse articulated objects using only a handheld RGB camera.

In summary, this work contributes:
\begin{enumerate}
    \item An extension of 3DGS with a differentiable mobility value per Gaussian that enables precise part segmentation using gradient-based optimization.
    \item A multi-stage optimization strategy that decouples reconstruction and articulation estimation, enhancing robustness and accuracy.
    \item Complementary formulations of geometric self-supervision for articulation estimation, eliminating the need for 3D supervision or laborious annotations.
    \item A challenging dataset and new metrics for comprehensive evaluation of articulated object reconstruction.
\end{enumerate}

\section{Related Work}\label{sec:related}

\subsection{Data-Driven Articulation Learning}
Estimating the pose and joint properties of articulated objects is crucial for robot manipulation and interaction~\cite{irshad2024neural, geng2023gapartnet, liu2022akb, geng2023partmanip}. Recent learning-based methods~\cite{yi2018deep, hu2017learning, wang2019shape2motion, heppert2023carto, jiang2022opd, liu2023building, li2020category, gadre2021act} infer articulation properties from point clouds via end-to-end training. For instance, Shape2Motion~\cite{wang2019shape2motion} analyzes motion parts from a single point cloud in a supervised setting, while ANCSH~\cite{li2020category} performs category-level pose estimation but requires class-specific models. RPM-Net~\cite{yan2020rpm} enhances generalization across categories for part segmentation and kinematic prediction, and DITTO~\cite{jiang2022ditto} predicts motion and geometry from 3D point cloud pairs without labels. However, these methods depend on costly 3D supervision and annotations. In contrast, our approach reconstructs accurate 3D geometry and detailed appearance, capturing articulation without 3D supervision or priors.

\subsection{Representations for Object Reconstruction}
Early 3D object reconstruction methods predicted point clouds, voxels, or meshes from partial observations~\cite{fan2017point, irshad2022centersnap, choy20163d}. Recent advances in implicit scene representations~\cite{park2019deepsdf, mescheder2019occupancy, irshad2022shapo, zakharov2024refine, mildenhall2021nerf, kerbl20233dgs} enable detailed geometry and appearance reconstruction via differentiable rendering~\cite{wang2021neus, takikawa2021neural, oechsle2021unisurf, guedon2023sugar}. While neural fields suffer from slow rendering, 3D Gaussian Splatting (3DGS)~\cite{kerbl20233dgs} overcomes this by using explicit 3D Gaussians. We leverage 3DGS for self-supervised articulated object reconstruction from posed RGB images, achieving fast, realistic synthesis of novel views and articulations in real time.

\subsection{Articulated Object Reconstruction}
Recent methods leverage differentiable 3D representations~\cite{park2019deepsdf, mildenhall2021nerf, kerbl20233dgs} to jointly reconstruct articulated objects and infer articulation parameters. Training-based approaches use synthetic 3D data to predict joint parameters and segment parts~\cite{mu2021sdf, jiang2022ditto, heppert2023carto, nie2022structure, hsu2023ditto, kawana2022unsupervised, wei2022self}. Self-supervised methods optimize shape, appearance, and articulation per scene without extensive training~\cite{liu2023paris, Liu_2023_CVPR, kerr2024robot, song2024reacto, weng2024neural, liu2025building, deng2024articulate}, with some addressing multi-part objects but requiring known part counts and single-level articulation structures~\cite{weng2024neural, deng2024articulate, liu2025building}. Other works enhance articulation estimation using large language or vision-language models~\cite{mandi2024real2code, le2024articulate}. In contrast, our self-supervised method reconstructs two-part articulated objects from RGB images across articulation states using 3DGS~\cite{kerbl20233dgs}. As the first to apply 3DGS to this task without 3D supervision or pre-trained priors, it robustly handles challenging cases and achieves real-time performance.

\begin{figure*}[htb]
     \centering
     \begin{subfigure}[t]{0.25\linewidth}
         \centering
         \captionsetup{justification=centering}
         \caption{
             \textbf{Stage 1: Per-state reconstruction.}\\
             (Section \ref{sec:method_stage_1})%
         }
         \includegraphics[height=3.7cm]{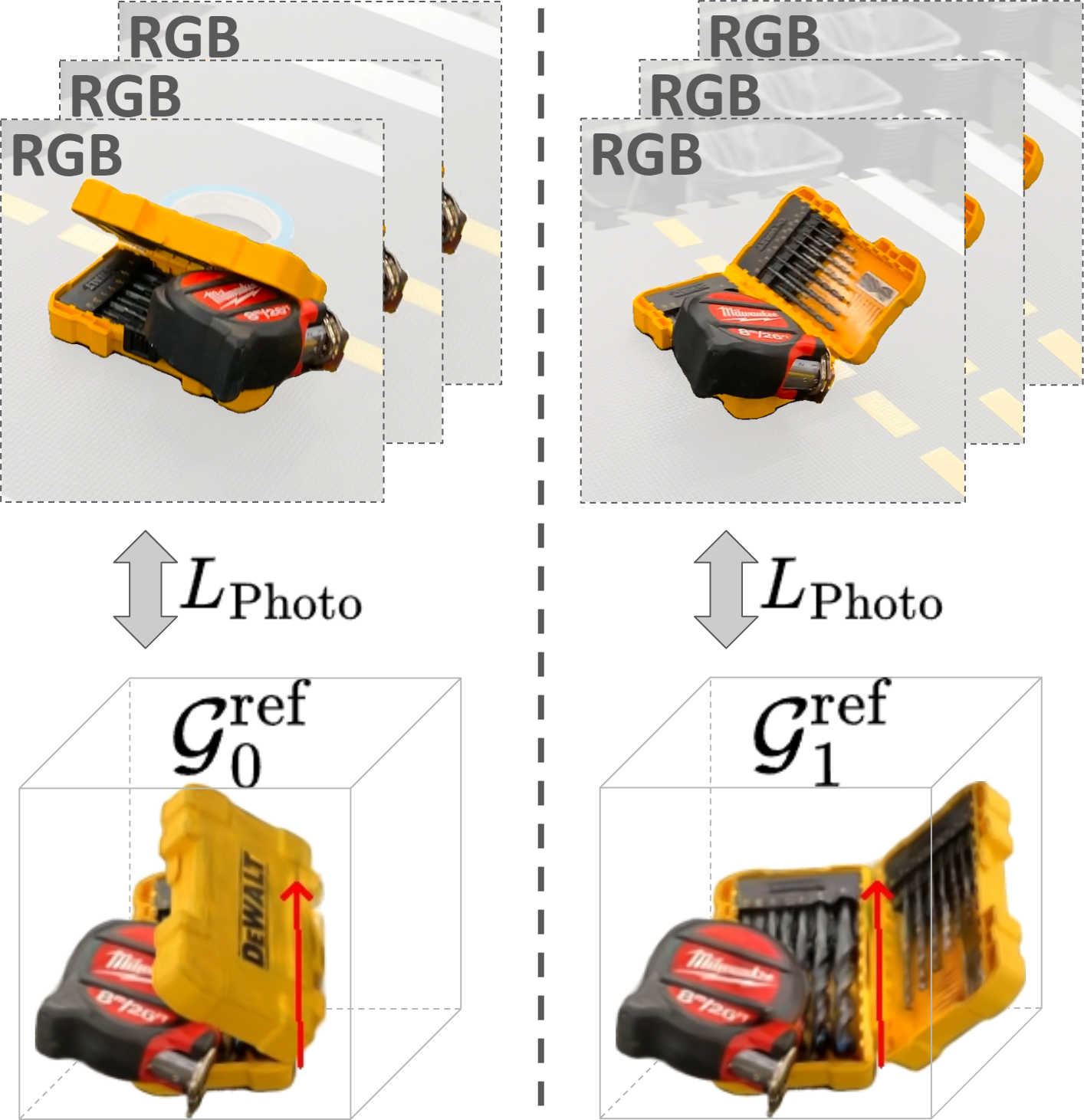}
         \label{fig:method_stage_1}
     \end{subfigure}%
     \hfill%
     \begin{subfigure}[t]{0.35\linewidth}
         \centering
         \captionsetup{justification=centering}%
         \caption{
             \textbf{Stage 2: Cross-static formulation\\ for mobility estimation.}\\ (Section \ref{sec:method_stage_2})%
         }
         \includegraphics[height=3.7cm]{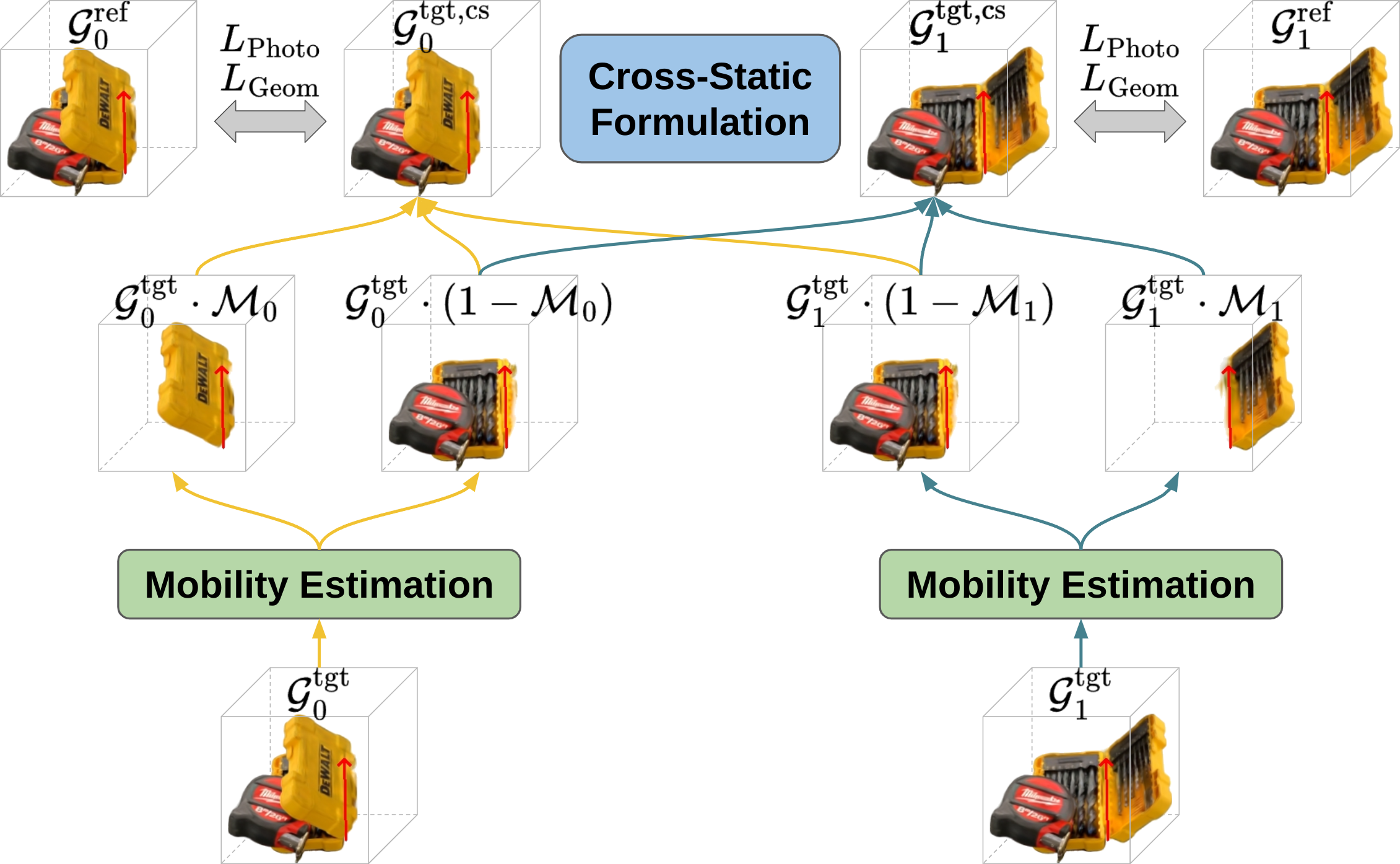}
         \label{fig:method_stage_2}
     \end{subfigure}%
     \hfill%
     \begin{subfigure}[t]{0.4\linewidth}
         \centering
         \captionsetup{justification=centering}%
         \caption{
             \textbf{Stage 3: Cross-mobile formulation for\\ articulation estimation and mobility refinement.}\\
             (Section \ref{sec:method_stage_3})
         }
         \includegraphics[height=3.7cm]{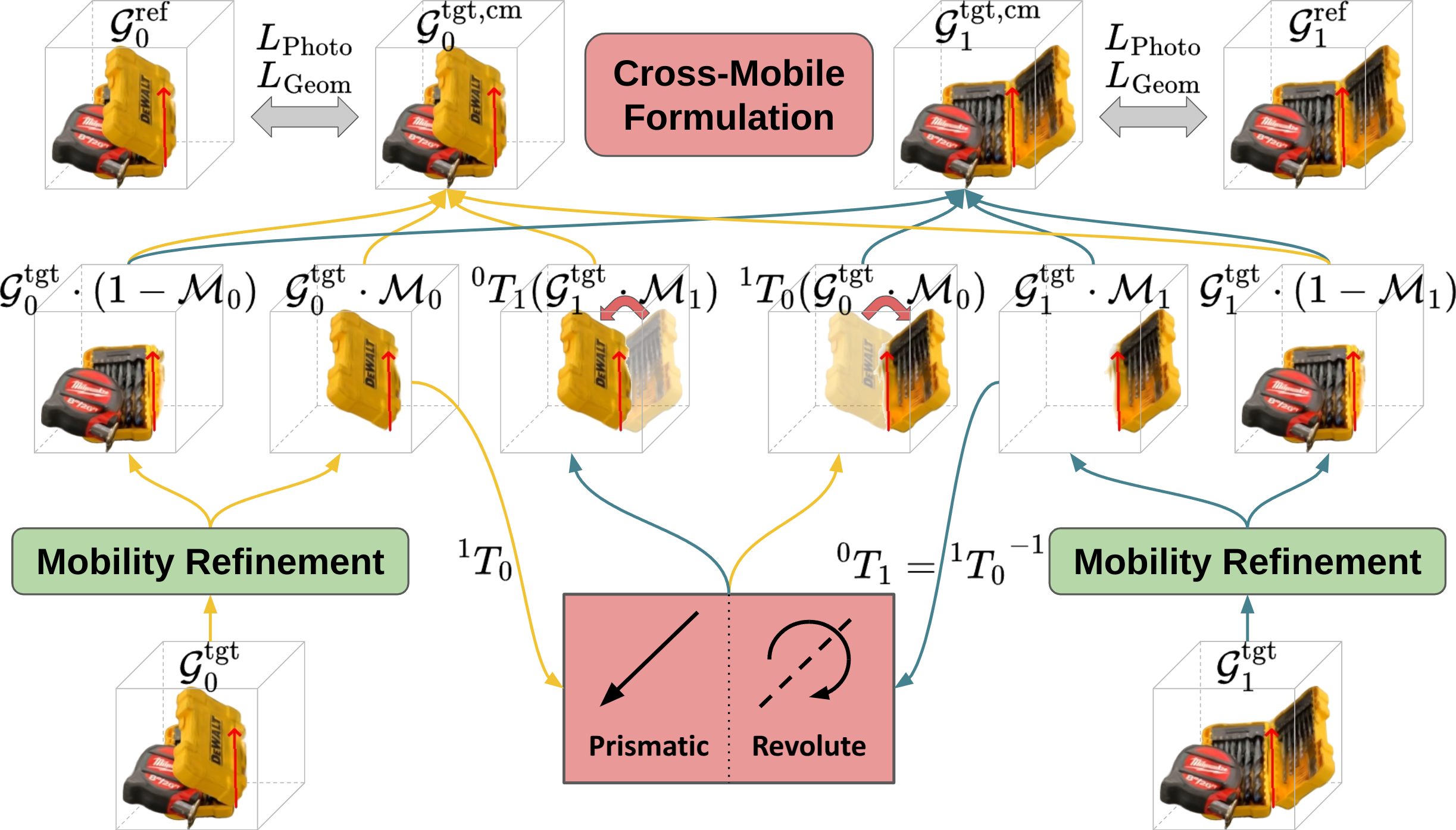}
         \label{fig:method_stage_3}
    \end{subfigure}
    \caption{\textbf{Methodology overview of \alg.} It consists of three decoupled stages for the purpose of stable optimization. \hyperref[fig:method_stage_1]{Stage~1} constructs a 3DGS for each state from posed RGB images using the photometric loss. \hyperref[fig:method_stage_2]{Stage~2} proposes cross-static formulation for mobility estimation. Intuitively, it combines static parts of both states and the mobile part of the desired state for the target Gaussians. \hyperref[fig:method_stage_3]{Stage~3} proposes cross-mobile formulation for articulation estimation. Intuitively, it combines static parts of both states, the mobile part of the desired state and the \emph{transformed} mobile part of the other state together for the target Gaussians.}
\end{figure*}

\section{Methodology}\label{sec:method}

\subsection{Overview}
Consider an arbitrary object composed of two rigid parts: a static \emph{parent} part, and a \emph{child} part that can move relative to its parent through either a revolute or prismatic articulation. Our objective is twofold:
\begin{enumerate*}[label=(\arabic*)]
    \item to reconstruct the articulated object at the part level; and
    \item to estimate its articulated motion.
\end{enumerate*}
Assuming a known articulation type (i.e., either revolute or prismatic), the input to our method consists of two sets of posed RGB images (i.e., images with known camera intrinsics and extrinsics), each capturing the articulated object at one end state of the motion.

Formally, let $l$ denote the articulation state label, where $l=0$ and $l=1$ correspond to the two end states of the observed articulation. For reconstruction, \alg uses observations $\mathcal{O}_l=\{(I_l^i,P_l^i,K_l^i)\}_{i=1}^{N_l},\ l\in\{0,1\}$, where $I_l^i$ is the $i$-th observed RGB image of the articulated object at state $l$, $P_l^i$ and $K_l^i$ denote its camera extrinsics and intrinsics respectively, and $N_l$ represents the number of data samples for state $l$. Note that $P_l^i$ is specified in a common world space for both states, while the articulated motion involves only one moving part w.r.t.\ the world space. \alg models a revolute articulation with its rotation axis $\mathbf{a}\ (\|\mathbf{a}\|=1)$, pivot $\mathbf{p}$, and rotation angle $\theta$, such that a point $\mathbf{x}$ (in the world space) on the mobile part at state $l=0$ will move to
\begin{equation}
    ^1T_0\mathbf{x}=R_{\mathbf{a},\theta}(\mathbf{x}-\mathbf{p})+\mathbf{p}
\end{equation}
at state $l=1$, where $R_{\mathbf{a},\theta}$ is the rotation induced by the axis-angle notation. A prismatic articulation is instead modeled by its translation axis $\mathbf{a}\ (\|\mathbf{a}\|=1)$ and distance $d$. The goal is to reconstruct the articulated object at the part level using a chosen representation while estimating the articulated motion $^1T_0$, ensuring that the renderings at each articulation state align consistently with the observations.

\alg extends the 3DGS representation for articulated objects, decoupling the goals of part-level reconstruction and articulation estimation across three stages:
\begin{enumerate*}[label=(\arabic*)]
    \item separate reconstructions for each articulation state (\cref{sec:method_stage_1}),
    \item mobility estimation using the cross-static formulation (\cref{sec:method_stage_2}), and
    \item articulation estimation and mobility refinement using the cross-mobile formulation (\cref{sec:method_stage_3}).
\end{enumerate*}
To facilitate the application of \alg to real-world objects, we leverage modern structure-from-motion and image segmentation techniques, developing a framework that enables general users to reconstruct articulated objects in their surroundings using only images captured by a handheld camera device (\cref{sec:method-real-world}).
\subsection{Decoupled Multi-Stage Optimization}
Jointly performing part-level reconstruction and articulation estimation on an articulated object is prone to local minima~\cite{liu2023paris,deng2024articulate}, making it preferable to strategically decouple the problem into multiple stages.
\subsubsection{Separate per-state reconstruction}\label{sec:method_stage_1}
In Stage~1, two 3DGS models are separately optimized following the standard procedure, one for each end state of the articulation. Specifically, apart from the attributes from original 3DGS, each Gaussian is additionally initialized with a persistent binary state label $l$, equally drawn from $\{0,1\}$. We denote the set of Gaussians representing state $l$ as $\mathcal{G}_l^\textrm{ref}$, where \emph{ref} emphasizes that $\mathcal{G}^\textrm{ref}$ is the reference reconstruction unaffected by the other state. Given a data sample observed at state $l$, we have the optimization:
\begin{subequations}
    \begin{gather}
        \min_{\mathcal{G}_l^\textrm{ref}}\;  \Delta_I(\hat{I}_l^{\textrm{ref},i},I_l^i),\\
        \hat{I}_l^{\textrm{ref},i}=R(\mathcal{G}_l^\textrm{ref},P_l^i,K_l^i),\ l\in\{0,1\},
    \end{gather}
\end{subequations}
where $R$ is the 3DGS rendering function, and $\Delta_I$ represents the photometric loss. For simplicity, the view index $i$ and camera parameters $P_l^i,K_l^i$ will be omitted from now on.

\subsubsection{Cross-static formulation for mobility estimation}\label{sec:method_stage_2}
The focus of Stage~2 is mobility estimation for each Gaussian. To keep $\mathcal{G}^\textrm{ref}$ dedicated to the single-state reconstruction, we first duplicate $\mathcal{G}_l^\textrm{ref}$ as $\mathcal{G}_l^\textrm{tgt}$ for both states (i.e., $l\in\{0,1\}$), which is intended as the target representation that will fulfill the goals of part-level reconstruction and articulation estimation. By design, $\mathcal{G}^\textrm{tgt}$ shares neither data storage nor gradient flow with $\mathcal{G}^\textrm{ref}$ after its creation.

For each Gaussian in $\mathcal{G}^\textrm{tgt}$, we further extend its set of attributes with a mobility value $m\in[0,1]$, initialized with $0.5$. By design, $m$ enables the break-down of a Gaussian to its static and mobile components, where each component inherits all the original Gaussian attributes except for the opacity $\sigma$. The static component has its opacity modified to $\sigma\cdot(1-m)$, and the mobile component to $\sigma\cdot{}m$. For simplicity, letting $\mathcal{M}$ be the set of mobilities for $\mathcal{G}$, we use the element-wise product $\mathcal{G}\cdot(1-\mathcal{M})$ to denote the static component of $\mathcal{G}$, and $\mathcal{G}\cdot\mathcal{M}$ for the mobile component.

To estimate the mobilities $\mathcal{M}$, we employ the intuition that the static components from both states should constitute the static part of the articulated object. Formally, we introduce the \emph{cross-static (cs) formulation}, where the static part of the articulated object is jointly represented as
\begin{equation}
    \mathcal{G}^\textrm{s}=\mathcal{G}_l^\textrm{tgt}\cdot(1-\mathcal{M}_l)\oplus\mathcal{G}_{1-l}^\textrm{tgt}\cdot(1-\mathcal{M}_{1-l}),\label{eq:s}
\end{equation}
where $\oplus$ denotes concatenation. For state $l$, the target representation thus becomes
\begin{equation}
    \mathcal{G}_l^{\textrm{tgt},\textrm{cs}} = \mathcal{G}^\textrm{s}\oplus\mathcal{G}_l^\textrm{tgt} \cdot \mathcal{M}_l.
\end{equation}
With this formulation, Stage~2 is further divided into two sub-stages as below.
    
\paragraph{Stage~2(a): Coarse mobility estimation via cross-static geometric consistency.} To measure the geometric distance between two Gaussian sets, we design a weighted version of the Chamfer distance. Specifically, let \(X = \{(x_i, w_{x_i})\}_{i=1}^M\) and \(Y = \{(y_j, w_{y_j})\}_{j=1}^N\) be two sets of point-weight pairs, the weighted Chamfer distance is then:
\begin{multline}
    \mathrm{Chamfer}(X,Y) = \sum_{\left(x_i, w_{x_i}\right) \in X} \tilde{w}_{x_i} \min_{\left(y_j, w_{y_j}\right) \in Y} \| x_i - y_j \|^2 \\
    + \sum_{\left(y_j, w_{y_j}\right) \in Y} \tilde{w}_{y_j} \min_{\left(x_i, w_{x_i}\right) \in X} \| x_i - y_j \|^2,
\end{multline}
where $\tilde{w}_{x_i} = \sfrac{w_{x_i}}{\sum_{(x_i, w_{x_i}) \in X} w_{x_i}}$ and $\tilde{w}_{y_j}$ are normalized weights. For each Gaussian, we use its mobility-modified opacity $\sigma'$ as the weight, essentially treating it as $\sigma'$ points overlapped at its mean position. The mobilities are then optimized by encouraging the geometric consistency formulated as follows:
\begin{subequations}
    \begin{gather}
        \min_\mathcal{M} \; \mathrm{CD}_0^\textrm{cs} + \mathrm{CD}_1^\textrm{cs} + \lambda_m^\textrm{geom}\|\mathcal{M}\|, \label{eq:s2-csgc}\\
        \mathrm{CD}_l^\textrm{cs} = \mathrm{Chamfer}(\mathcal{G}_l^{\textrm{tgt},\textrm{cs}},\mathcal{G}_l^\textrm{ref}),
    \end{gather}
\end{subequations}
where $\textrm{Chamfer}(\cdot)$ denotes the weighted Chamfer distance, and $\lambda_m^\textrm{geom}\|\mathcal{M}\|$ is the regularization term that encourages
smaller mobilities. Note how $\mathcal{M}$ affects the weighted Chamfer distance by modifying the opacities, and that $\mathcal{M}=1$ is a trivial solution without the regularization. 
Without photometric supervision, the mobilities obtained from Eqn.~\ref{eq:s2-csgc} are generally noisy. However, being relatively fast (taking only tens of seconds), they still serve as a good initialization for the next sub-stage.

\paragraph{Stage~2(b): Joint mobility and Gaussian optimization via cross-static rendering.} To more accurately estimate the mobilities while jointly refining the full Gaussian attributes, cross-static rendering is performed as follows:
\begin{subequations}
\small
    \begin{gather}
        \min_{\mathcal{G}^\textrm{tgt},\mathcal{M}_{1-l}}\Delta_I(\hat{I}_l^\textrm{cs},I_l)+\lambda_m^\textrm{photo}\|\mathcal{M}_{1-l}\|,\\
        \hat{I}_l^\textrm{cs} = R(\mathcal{G}_l^{\textrm{tgt},\textrm{cs}}),
    \end{gather}
\end{subequations}
where $\lambda_m^\textrm{photo}\|\mathcal{M}_{1-l}\|$ is the mobility regularization term similar to that in Eqn.~\ref{eq:s2-csgc}. 

\subsubsection{Cross-mobile formulation for articulation estimation and mobility refinement}\label{sec:method_stage_3}
The focus of Stage~3 is to estimate the articulation parameters shared by the mobile components of all Gaussians. To this end, we employ the intuition that the mobile components from the two end states are related through the articulated motion. Formally, we introduce the \emph{cross-mobile (cm) formulation}, where the mobile part of the articulated object at state $l$ is jointly represented as
\begin{equation}
    \mathcal{G}_l^\textrm{m}=\mathcal{G}_l^\textrm{tgt}\cdot\mathcal{M}_l\oplus{}^lT_{1-l}(\mathcal{G}_{1-l}^\textrm{tgt}\cdot\mathcal{M}_{1-l}),
\end{equation}
where ${}^lT_{1-l}(\mathcal{G})$ denotes the transformation of Gaussians $\mathcal{G}$ according to the articulated motion from state $1-l$ to $l$. For state $l$, the target representation thus becomes
\begin{equation}
    \mathcal{G}_l^{\textrm{tgt},\textrm{cm}}=\mathcal{G}^\textrm{s}\oplus\mathcal{G}_l^\textrm{m},
\end{equation}
where $\mathcal{G}^\textrm{s}$ is defined in Eqn.~\ref{eq:s}. With this formulation, Stage~3 is further divided into three sub-stages as below.

\begin{figure}[!tb]
    \centering
    \begin{subfigure}{0.48\linewidth}
        \centering
        \includegraphics[width=\linewidth]{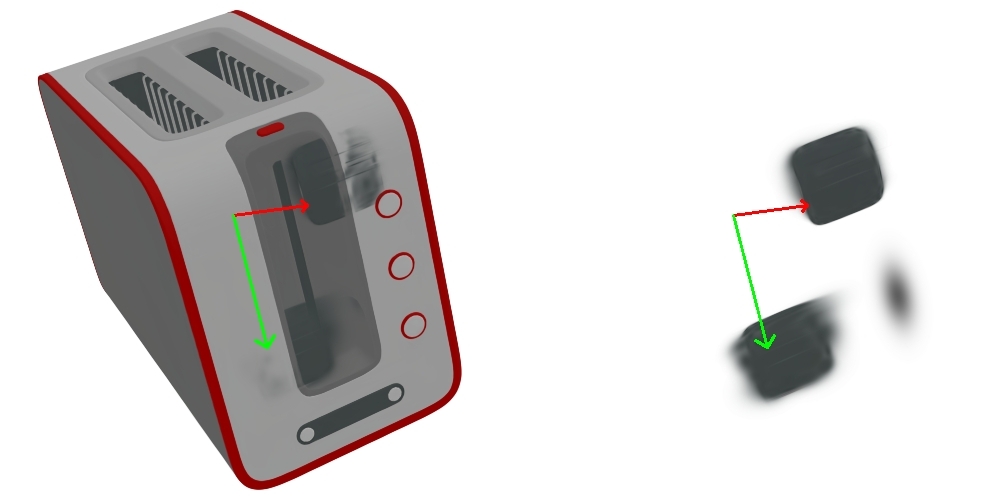}
        \caption{Failure with cross-mobile geometric consistency.}
        \label{fig:cm-failure}
    \end{subfigure}
    \hfill
    \begin{subfigure}{0.48\linewidth}
        \centering
        \includegraphics[width=\linewidth]{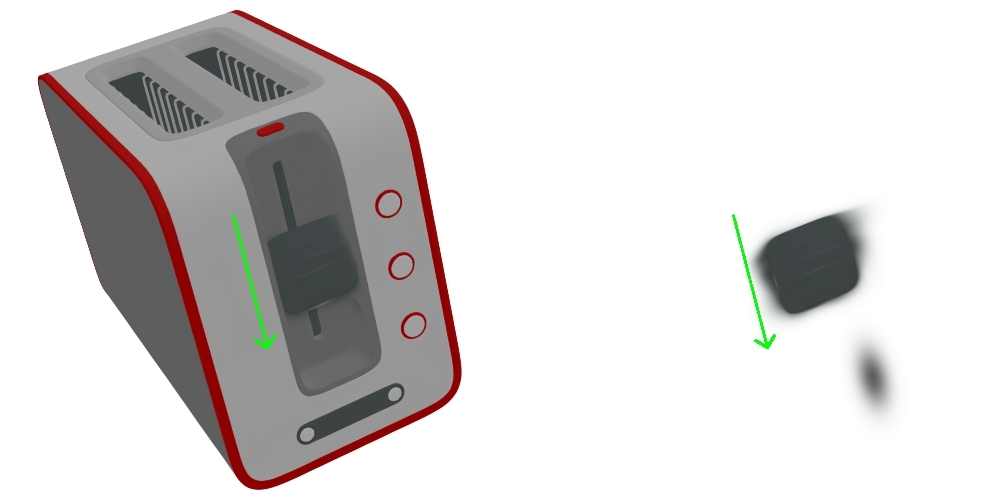}
        \caption{Success with mobile-only geometric consistency.}
        \label{fig:m-success}
    \end{subfigure}
    \caption{103549-Toaster has a flat slider as its mobile part, similar in geometric curvature to the static component. This similarity makes the nearest-neighbor data association error-prone when computing the Chamfer distance for cross-mobile geometric consistency. On the other hand, mobile-only geometric consistency focuses only on the mobile parts, successfully handling the case.}
    \label{fig:toaster}
\end{figure}

\paragraph{Stage 3(a): Coarse articulation estimation via geometric consistency.} Similar to Eqn.~\ref{eq:s2-csgc}, weighted Chamfer distance is used for computing the cross-mobile geometric consistency, which can be optimized over both mobilities and articulation parameters as follows:
\begin{subequations}
    \begin{gather}
        \min_{\mathcal{M},T} \; \mathrm{CD}_0^{\textrm{cm}}+\mathrm{CD}_1^{\textrm{cm}},\label{eq:cmgc}\\
        \mathrm{CD}_l^{\textrm{cm}} = \mathrm{Chamfer}(\mathcal{G}_l^{\textrm{tgt},\textrm{cm}},\mathcal{G}_l^{\textrm{ref}}).
    \end{gather}
\end{subequations}
However, we notice that this formulation is still susceptible to local minima if the mobile part is too geometrically insignificant. We qualitatively show one such failure in \cref{fig:cm-failure}. To remedy this, we further propose mobile-only geometric consistency, which concerns only the mobile components of both states as follows:
\begin{equation}
    \mathrm{CD}^\textrm{m}=\mathrm{Chamfer}(\mathcal{G}_l^{\textrm{tgt}}\cdot\mathcal{M}_l,{}^lT_{1-l}(\mathcal{G}_{1-l}^\textrm{tgt}\cdot\mathcal{M}_{1-l})).
\end{equation}

\begin{figure}[!tb]
    \centering
    \begin{subfigure}{0.48\linewidth}
        \centering
        \includegraphics[width=\linewidth]{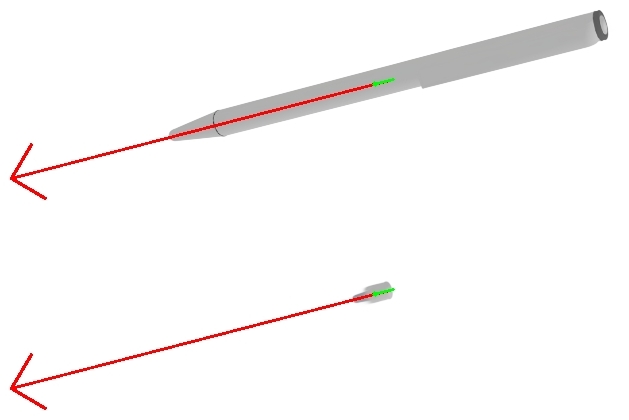}
        \caption{Failure with mobile-only geometric consistency.}
        \label{fig:m-failure}
    \end{subfigure}
    \hfill
    \begin{subfigure}{0.48\linewidth}
        \centering
        \includegraphics[width=\linewidth]{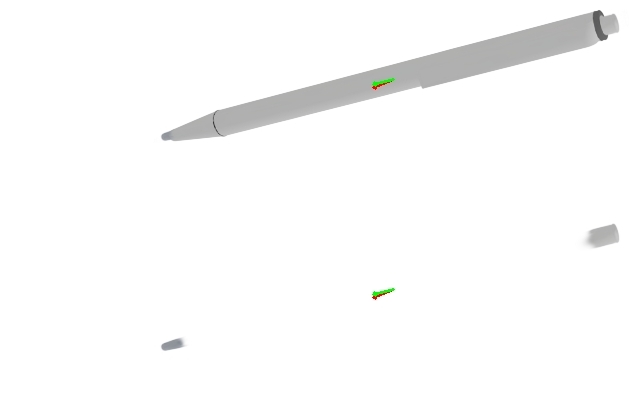}
        \caption{Success with cross-mobile geometric consistency.}
        \label{fig:cm-success}
    \end{subfigure}
    \caption{101713-Pen has both ends of a pen as the mobile part, with only one end visible from any given observation. Thus, the reconstructed mobile components from the two states essentially represent distinct parts, violating the intuition for mobile-only geometric consistency. On the other hand, cross-mobile geometric consistency considers the joint representation from both states as a whole, successfully estimating the articulation.}
    \label{fig:pen}
\end{figure}

While successfully addressing these scenarios, mobile-only geometric consistency still falls short in other circumstances, especially when the mobile components in the two states exhibit large discrepancies. We qualitatively show one such failure in \cref{fig:m-failure}. To take advantage of both formulations and to facilitate robustness in the inherently non-convex optimization of articulation parameters, we propose the following practical scheme:
\begin{enumerate}
    \item With $K^\textrm{m}$ randomized tries: articulation estimation via
    \begin{equation}
        T^m=\arg\min_T\mathrm{CD}^\textrm{m}.
    \end{equation}
    \item With $K^\textrm{cm}$ randomized tries, plus another initialized with $T^\textrm{m}$: articulation estimation via
    \begin{equation}
        T^\textrm{cm}=\arg\min_T \; \mathrm{CD}_0^\textrm{cm}+\mathrm{CD}_1^\textrm{cm}.
    \end{equation}
    \item Final run initialized with $T^\textrm{cm}$: joint articulation estimation and mobility refinement via Eqn.~\ref{eq:cmgc}.
\end{enumerate}

\paragraph{Stage 3(b): Joint articulation, mobility, and Gaussian optimization via cross-mobile rendering.} Like in Stage 2(b), we perform a full joint optimization over articulation parameters, mobilities, and Gaussians utilizing photometric supervision via cross-mobile rendering as follows:
\begin{subequations}
    \begin{gather}
        \min_{\mathcal{G}^\textrm{tgt},\mathcal{M}_{1-l},{}^lT_{1-l}} \;  \Delta_I(\hat{I}_l^{\textrm{cm}},I_l),\\
        \hat{I}_l^{\textrm{cm}} = R(\mathcal{G}_l^{\textrm{tgt},\textrm{cm}}),\ l\in\{0,1\}.
    \end{gather}
\end{subequations}

\paragraph{Stage 3(c): Mobility correction via cross-mobile geometric consistency.} Stage 3(b) solely relies on photometric supervision, which is limited to the training views. On rare occasions, a Gaussian may be mistakenly estimated as mobile if the articulated motion moves it out of sight from most views, effectively losing the supervision. On the other hand, geometric consistency is not affected by sight limitations, which we leverage for mobility correction as follows:
\begin{equation}
    \min_\mathcal{M} \; \mathrm{CD}_0^\textrm{cm}+\mathrm{CD}_1^\textrm{cm}.
\end{equation}

\subsection{Real-World Application}\label{sec:method-real-world}
For an articulated object to be reconstructed in the real world, we first collect two sets of RGB images, one for each articulation state, by imaging the object from a surrounding hemisphere. We then preprocess the data to determine the posed images that serve as the input to \alg. This involves using SAM 2~\cite{ravi2024sam} to perform foreground-background segmentation, which also removes dynamic contents of the scene. We then perform structure-from-motion (SfM) to determine camera poses using COLMAP~\cite{schonberger2016structure} with SuperPoint~\cite{detone2018superpoint} descriptors and SuperGlue~\cite{sarlin2020superglue} matching on the segmented backgrounds from both sets of images to construct a joint coordinate frame for the object in both states. Once we obtain the joint coordinate frame and the foreground target object is localized, we run \alg to reconstruct the articulated objects.

\section{Experiments}\label{sec:experiments}

\begin{table*}[htb]
    \resizebox{\linewidth}{!}{
        \begin{tabular}{llcS[table-format=3.2(4)]S[table-format=1.2(3)]S[table-format=3.2(4)]S[table-format=2.3(3)]S[table-format=2.2(2.2)]S[table-format=3.2(2.2)]S[table-format=2.2(1.2)]}
            \toprule
            Type & Method & \makecell{Success\\ Rate} $\big\uparrow$ & {\makecell{$\textrm{err}_a$ \\ ($\times10^{-2}$ DEG)} $\big\downarrow$} & {\makecell{$\textrm{err}_p$ \\ $(\times10^{-3})$} $\big\downarrow$} & {\makecell{$\textrm{err}_r$ \\ ($\times10^{-2}$ DEG)} $\big\downarrow$} & {\makecell{$\textrm{err}_t$ \\ $(\times10^{-3})$} $\big\downarrow$} & {\makecell{$\textrm{CD}_s$ \\ $(\times10^{-3})$} $\big\downarrow$} & {\makecell{$\textrm{CD}_m$ \\ $(\times10^{-3})$} $\big\downarrow$} & {\makecell{$\textrm{CD}_w$ \\ $(\times10^{-3})$} $\big\downarrow$} \\
            \midrule
            \multirow{3}{*}{Revolute} & PARIS & $40.0\%$ & 121.90 (68.45) & 5.27 (4.55) & 127.80 (55.12) & {N/A} & 3.48 (0.39) & 68.44 (13.03) & 13.11 (1.16) \\
            & DTA$^\dagger$ & $98.8\%$ & 12.32 (4.16) & 1.86 (1.69) & 19.41 (16.28) & {N/A} & 2.20 (0.49) & 1.27 (1.87) & 1.73 (0.03) \\
            & \alg & $100.0\%$ & 3.70 (0.38) & 0.40 (0.07) & 4.85 (0.33) & {N/A} & 4.08 (0.32) & 1.06 (0.78) & 3.59 (0.21) \\
            \dashrule{1-10}
            \multirow{3}{*}{Prismatic} & PARIS & $50.0\%$ & 27.97 (13.09) & {N/A} & {N/A} & 4.28 (3.02) & 9.21 (1.94) & 151.78 (35.00) & 7.99 (0.49) \\
            & DTA$^\dagger$ & $100.0\%$ & 16.26 (3.54) & {N/A} & {N/A} & 1.15 (0.14) & 2.69 (0.04) & 15.74 (0.28) & 2.22 (0.03) \\
            & \alg & $95.0\%$ & 2.36 (0.44) & {N/A} & {N/A} & 0.33 (0.04) & 6.02 (0.20) & 16.77 (1.14) & 3.69 (0.14) \\
            \dashrule{1-10}
            \multirow{3}{*}{Overall} & PARIS & $42.0\%$ & 103.12 (57.38) & 5.27 (4.55) & 127.80 (55.12) & 4.28 (3.02) & 4.63 (0.70) & 85.11 (17.42) & 12.09 (1.03) \\
            & DTA$^\dagger$ & $99.0\%$ & 13.11 (4.04) & 1.86 (1.69) & 19.41 (16.28) & 1.15 (0.14) & 2.30 (0.40) & 4.16 (1.55) & 1.83 (0.03) \\
            & \alg & $99.0\%$ & 3.44 (0.39) & 0.40 (0.07) & 4.85 (0.33) & 0.33 (0.04) & 4.47 (0.30) & 4.20 (0.85) & 3.61 (0.19) \\
            \bottomrule
        \end{tabular}
    }
    \caption{PARIS-PMS Articulation and Mesh Reconstruction Metrics. $^\dagger$DTA requires ground-truth depth.}
    \label{tab:paris-pms-articulation-mean}
\end{table*}

\begin{table}[htb]
    \resizebox{\linewidth}{!}{
        \begin{tabular}{llccc}
            \toprule
            Type & Method & PSNR $\uparrow$ & Depth MAE $\downarrow$ & mIoU $\uparrow$ \\
            \midrule
            \multirow{3}{*}{Revolute} & PARIS & $32.21$ & $0.093$ & $0.955$ \\
            & DTA$^\dagger$ & N/A & $0.031$ & $0.941$ \\
            & \alg & $43.53$ & $0.039$ & $0.974$ \\
            \dashrule{1-5}
            \multirow{3}{*}{Prismatic} & PARIS & $33.75$ & $0.108$ & $0.902$ \\
            & DTA$^\dagger$ & N/A & $0.066$ & $0.844$ \\
            & \alg & $44.38$ & $0.025$ & $0.890$ \\
            \dashrule{1-5}
            \multirow{3}{*}{Overall} & PARIS & $32.52$ & $0.096$ & $0.942$ \\
            & DTA$^\dagger$ & N/A & $0.038$ & $0.922$ \\
            & \alg & $43.70$ & $0.036$ & $0.957$ \\
            \bottomrule
        \end{tabular}
    }
    \caption{PARIS-PMS Novel View Synthesis Metrics. Results correspond to average over successful runs (see Tab.~\ref{tab:paris-pms-articulation-mean} for success rates). $^\dagger$DTA requires ground-truth depth.}
    \label{tab:paris-pms-synthesis-mean}
\end{table}

\subsection{Datasets}\label{sec:datasts}
\paragraph{PARIS PartNet-Mobility Subset.}
PartNet-Mobility is a large-scale dataset that provides simulatable 3D object models with part-level mobility~\cite{Xiang_2020_SAPIEN,Mo_2019_CVPR,chang2015shapenet}, from which PARIS~\cite{liu2023paris} selects 10 instances for experiments, 8 being revolute and 2 being prismatic. We refer to this dataset as PARIS-PMS. For each articulation state, PARIS provides 100 calibrated object-centric views for training and 50 for testing, sampled from the upper hemisphere. However, the released dataset lacks ground-truth depth and part segmentation maps. To address this, we follow their data generation procedure and augment PARIS-PMS with the necessary ground-truth data for depth and segmentation evaluations. Still, no test views for intermediate articulation states are provided, limiting quantitative evaluation of novel articulation synthesis.
\paragraph{\alg PartNet-Mobility Subset.}
We curate an additional articulated object dataset from PartNet-Mobility, dubbed \alg-PMS, to enable a more comprehensive evaluation, including the quantitative assessment of novel articulation synthesis. Our dataset comprises 22 object instances from categories not included in PARIS-PMS, with 12 revolute and 10 prismatic objects. For the test split, we also generate ground-truth depth and part segmentation maps to facilitate relevant evaluations. Crucially, while PARIS-PMS restricts articulation states to binary values, \alg-PMS uniformly samples states from $[-0.1, 1.1]$, demanding accurate estimation of articulation, part segmentation, and object reconstruction for correct view synthesis. Our experiments reveal that \alg-PMS poses a greater challenge than PARIS-PMS—the latter being almost perfectly solved by our method.

\begin{table}[!tb]
    \centering
    \newcolumntype{C}[1]{>{\centering\arraybackslash}m{#1}}
    \begin{tabular}{C{0.2\linewidth}C{0.2\linewidth}C{0.2\linewidth}C{0.2\linewidth}}
        State 0 & Static Part &  Mobile Part & State 1 \\
        \midrule
        \multicolumn{4}{c}{\includegraphics[width=1.0\linewidth]{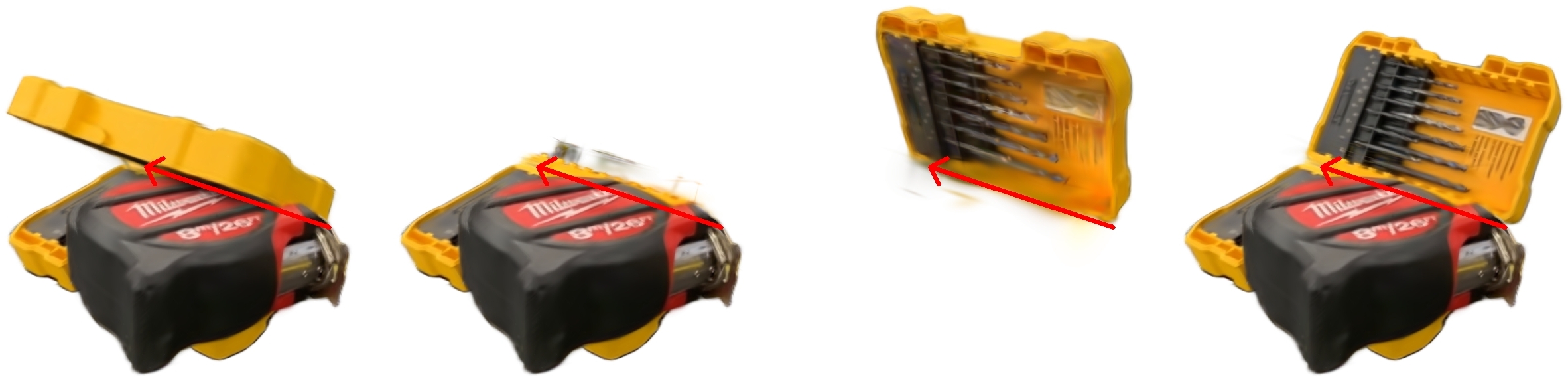}} \\
        \multicolumn{4}{c}{\includegraphics[width=1.0\linewidth]{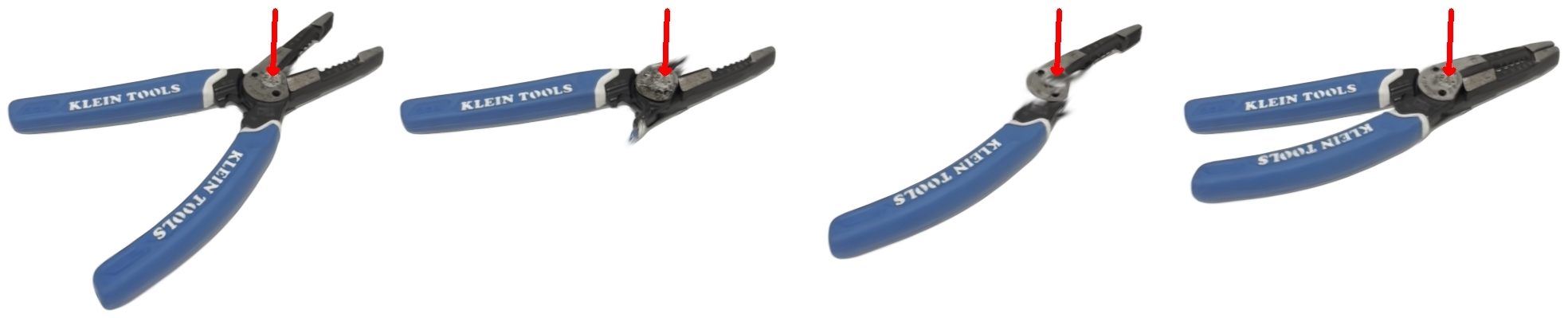}} \\
        \multicolumn{4}{c}{\includegraphics[width=1.0\linewidth]{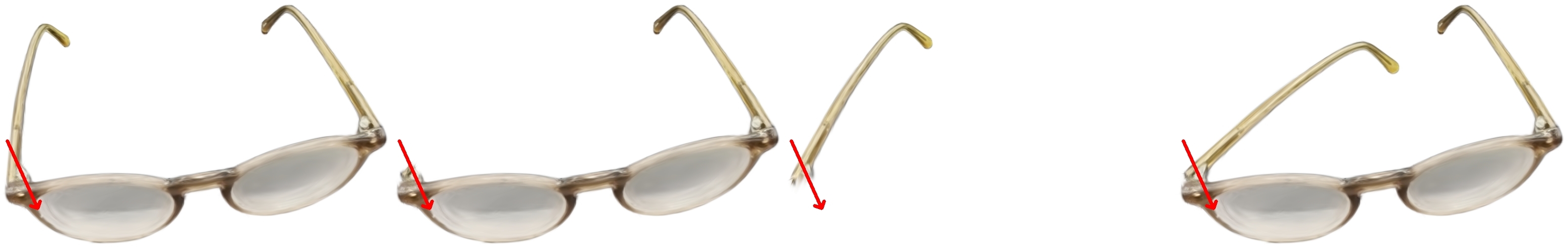}} \\
    \end{tabular}
    \caption{Real-World Articulated Object Reconstructions}
    \label{tab:qualitative-real}
\end{table}

\paragraph{Real-World Dataset.}
To show the efficacy of \alg on real-world usage, we collect a dataset consisting of 7 real-world objects with 9 articulations in total, including common objects like monitor, cabinet, glasses, etc. All images are captured by a hand-held phone camera. Camera parameters are calibrated following strategies detailed in \cref{sec:method-real-world}. We qualitatively evaluate \alg on this dataset.

\begin{table}[htb]
    \centering
    \resizebox{\linewidth}{!}{
    \setlength{\tabcolsep}{3pt}
    \begin{tabular}{llccccc}
        \toprule
        Type & Method & \makecell{Success\\ Rate} $\big\uparrow$ & {\makecell{$\textrm{err}_a$ \\ ($\times10^{-2}$ DEG)} $\big\downarrow$} & {\makecell{$\textrm{err}_p$ \\ $(\times10^{-3})$} $\big\downarrow$} & {\makecell{$\textrm{err}_r$ \\ ($\times10^{-2}$ DEG)} $\big\downarrow$} & {\makecell{$\textrm{err}_t$ \\ $(\times10^{-3})$} $\big\downarrow$} \\
        \midrule
        \multirow{7}{*}{\rotatebox{90}{Revolute}} & PARIS & $18.3\%$ & \nsd{167.34}{42.47} & \nsd{9.36}{3.73} & \nsd{270.37}{91.19} & {N/A} \\
        & DTA$^\dagger$ & $33.3\%$ & \nsd{\hphantom{1}32.76}{2.38\hphantom{1}} & \nsd{1.28}{0.39} & \nsd{\hphantom{1}40.80}{4.66\hphantom{1}} & {N/A} \\
        & \alg-2a & $78.5\%$ & \nsd{\hphantom{11}2.91}{0.55\hphantom{1}} & \nsd{2.53}{2.18} & \nsd{\hphantom{1}52.06}{32.00} & {N/A} \\
        & \alg-2b & $67.0\%$ & \nsd{\hphantom{11}4.62}{1.69\hphantom{1}} & \nsd{3.92}{3.07} & \nsd{131.99}{64.45} & {N/A} \\
        & \alg-3a & $\hphantom{0}0.8\%$ & \nsd{296.61}{0.00\hphantom{1}} & \nsd{2.83}{0.00} & \nsd{427.10}{0.00\hphantom{1}} & {N/A} \\
        & \alg-3b & $76.7\%$ & \nsd{\hphantom{1}84.13}{19.13} & \nsd{5.11}{3.03} & \nsd{206.98}{56.30} & {N/A} \\
        & \alg & $77.5\%$ & \nsd{\hphantom{1}20.76}{0.81\hphantom{1}} & \nsd{3.93}{2.41} & \nsd{\hphantom{1}78.31}{31.06} & {N/A} \\
        \dashrule{1-7}
        \multirow{7}{*}{\rotatebox{90}{Prismatic}} & PARIS & $39.0\%$ & \nsd{\hphantom{1}49.49}{18.92} & {N/A} & {N/A} & \nsd{14.96}{5.51} \\
        & DTA$^\dagger$ & $90.0\%$ & \nsd{102.77}{12.48} & {N/A} & {N/A} & \nsd{\hphantom{1}9.25}{0.73} \\
        & \alg-2a & $90.0\%$ & \nsd{\hphantom{1}25.88}{23.25} & {N/A} & {N/A} & \nsd{\hphantom{1}1.50}{1.63} \\
        & \alg-2b & $85.7\%$ & \nsd{\hphantom{1}15.63}{20.40} & {N/A} & {N/A} & \nsd{\hphantom{1}0.73}{0.58} \\
        & \alg-3a & $13.0\%$ & \nsd{\hphantom{1}15.17}{0.59\hphantom{1}} & {N/A} & {N/A} & \nsd{\hphantom{1}2.26}{0.04} \\
        & \alg-3b & $86.0\%$ & \nsd{\hphantom{1}61.00}{19.76} & {N/A} & {N/A} & \nsd{\hphantom{1}9.43}{2.37} \\
        & \alg & $95.0\%$ & \nsd{\hphantom{1}25.72}{17.63} & {N/A} & {N/A} & \nsd{\hphantom{1}1.04}{0.32} \\
        \dashrule{1-7}
        \multirow{7}{*}{\rotatebox{90}{Overall}} & PARIS & $27.7\%$ & \nsd{103.88}{29.79} & \nsd{9.36}{3.73} & \nsd{270.37}{91.19} & \nsd{14.96}{5.51} \\
        & DTA$^\dagger$ & $59.1\%$ & \nsd{\hphantom{1}77.76}{8.87\hphantom{1}} & \nsd{1.28}{0.39} & \nsd{\hphantom{1}40.80}{4.66\hphantom{1}} & \nsd{\hphantom{1}9.25}{0.73} \\
        & \alg-2a & $83.7\%$ & \nsd{\hphantom{1}13.85}{11.36} & \nsd{2.53}{2.18} & \nsd{\hphantom{1}52.06}{32.00} & \nsd{\hphantom{1}1.50}{1.63} \\
        & \alg-2b & $74.7\%$ & \nsd{\hphantom{11}9.34}{9.71\hphantom{1}} & \nsd{3.92}{3.07} & \nsd{131.99}{64.45} & \nsd{\hphantom{1}0.73}{0.58} \\
        & \alg-3a & $\hphantom{0}6.4\%$ & \nsd{\hphantom{1}85.53}{0.44\hphantom{1}} & \nsd{2.83}{0.00} & \nsd{427.10}{0.00\hphantom{1}} & \nsd{\hphantom{1}2.26}{0.04} \\
        & \alg-3b & $80.9\%$ & \nsd{\hphantom{1}73.62}{19.42} & \nsd{5.11}{3.03} & \nsd{206.98}{56.30} & \nsd{\hphantom{1}9.43}{2.37} \\
        & \alg & $85.5\%$ & \nsd{\hphantom{1}23.02}{8.46\hphantom{1}} & \nsd{3.93}{2.41} & \nsd{\hphantom{1}78.31}{31.06} & \nsd{\hphantom{1}1.04}{0.32} \\
        \bottomrule
    \end{tabular}
    }
    \caption{\alg-PMS Articulation Metrics. $^\dagger$DTA requires ground-truth depth.}
    \label{tab:splart-pms-articulation}
\end{table}

\begin{table}[htb]
    \resizebox{\linewidth}{!}{
        \begin{tabular}{llccc}
            \toprule
            Type & Method & PSNR $\uparrow$ & Depth MAE $\downarrow$ & mIoU $\uparrow$ \\
            \midrule
            \multirow{3}{*}{Revolute} & PARIS & $32.75$ & $0.125$ & $0.897$ \\
            & DTA$^\dagger$ & N/A & $0.053$ & $0.886$ \\
            & SplArt & $36.38$ & $0.032$ & $0.897$ \\
            \dashrule{1-5}
            \multirow{3}{*}{Prismatic} & PARIS & $32.30$ & $0.124$ & $0.902$ \\
            & DTA$^\dagger$ & N/A & $0.060$ & $0.857$ \\
            & SplArt & $37.77$ & $0.033$ & $0.938$ \\
            \dashrule{1-5}
            \multirow{3}{*}{Overall} & PARIS & $32.51$ & $0.125$ & $0.900$ \\
            & DTA$^\dagger$ & N/A & $0.058$ & $0.867$ \\
            & SplArt & $37.01$ & $0.032$ & $0.916$ \\
            \bottomrule
        \end{tabular}
    }
    \caption{\alg-PMS Novel View and Articulation Synthesis Metrics. Results correspond to average over successful runs (see Tab.~\ref{tab:splart-pms-articulation} for success rates). $^\dagger$DTA requires ground-truth depth.}
    \label{tab:splart-pms-synthesis}
\end{table}

\begin{table*}[htb]
\footnotesize
    \centering
    \setlength{\tabcolsep}{5pt}
    \begin{tabular}{ccc}
        \toprule
        \multirow{2}{*}{Input States} & Color & Part Segmentation \\
        \cmidrule(lr){2-3}
        & \hspace{0.5em} PARIS \hspace{2em} \alg \hspace{1em} Ground-truth & \hspace{0.5em} PARIS \hspace{2em} \alg \hspace{1em} Ground-truth \\
        \midrule
        \multicolumn{1}{c|}{\includegraphics[width=0.22\textwidth, align=c]{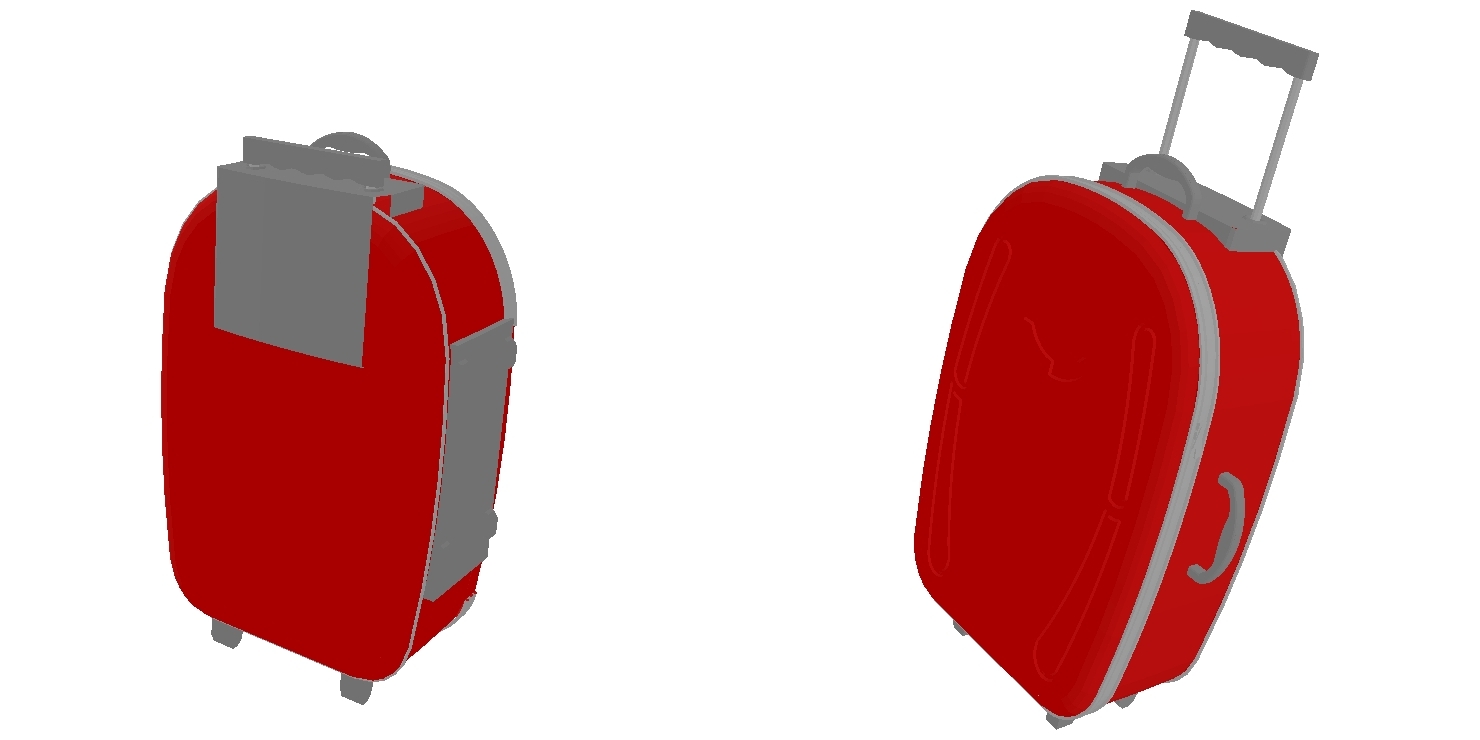}} & \includegraphics[width=0.35\textwidth, align=c]{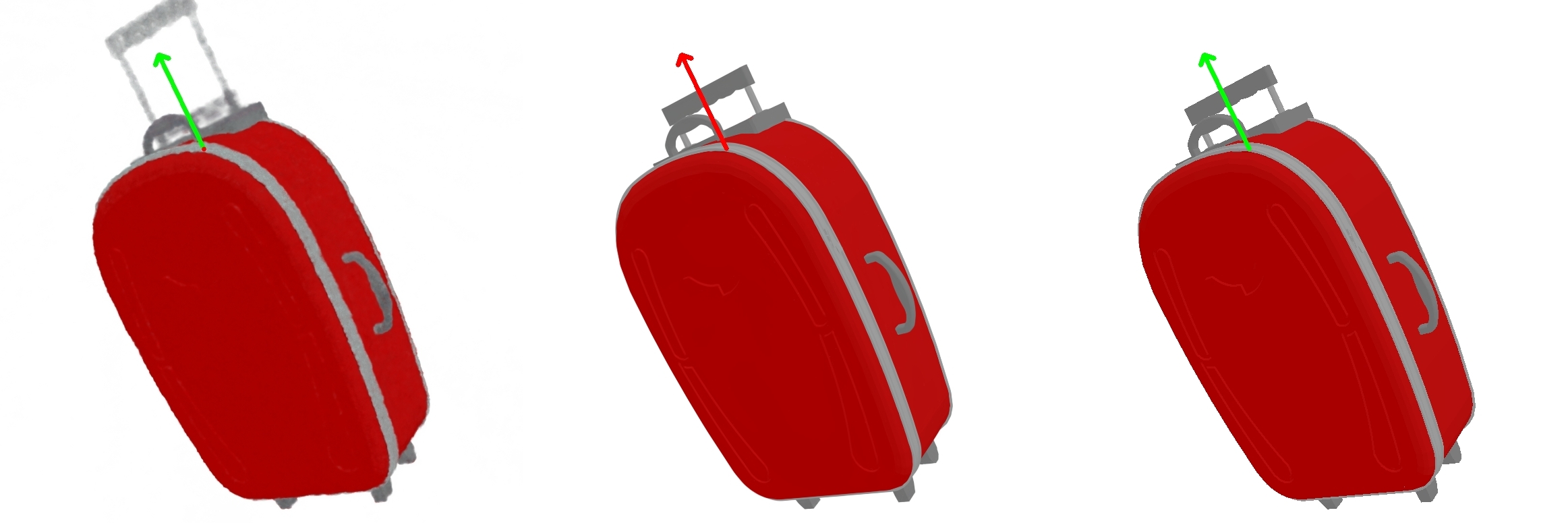} & \includegraphics[width=0.35\textwidth, align=c]{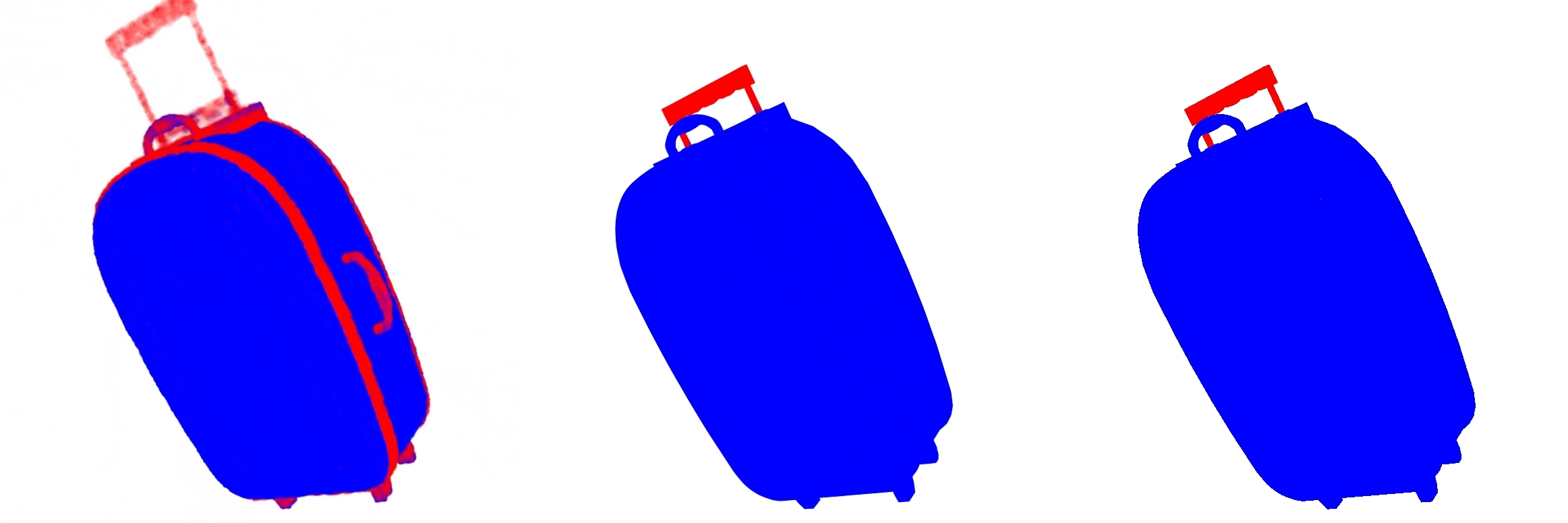} \\
        \multicolumn{1}{c|}{\includegraphics[width=0.22\textwidth, align=c]{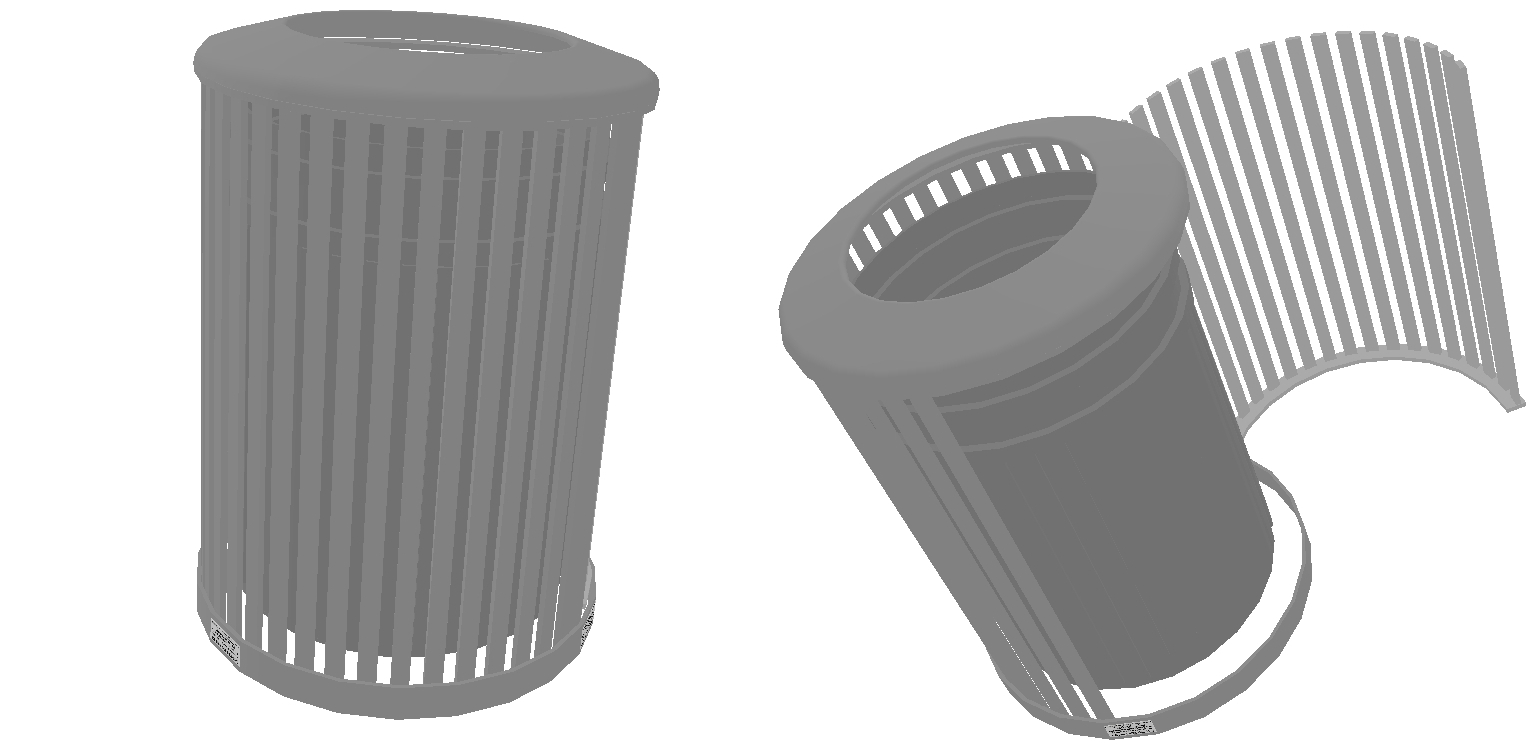}} & \includegraphics[width=0.35\textwidth, align=c]{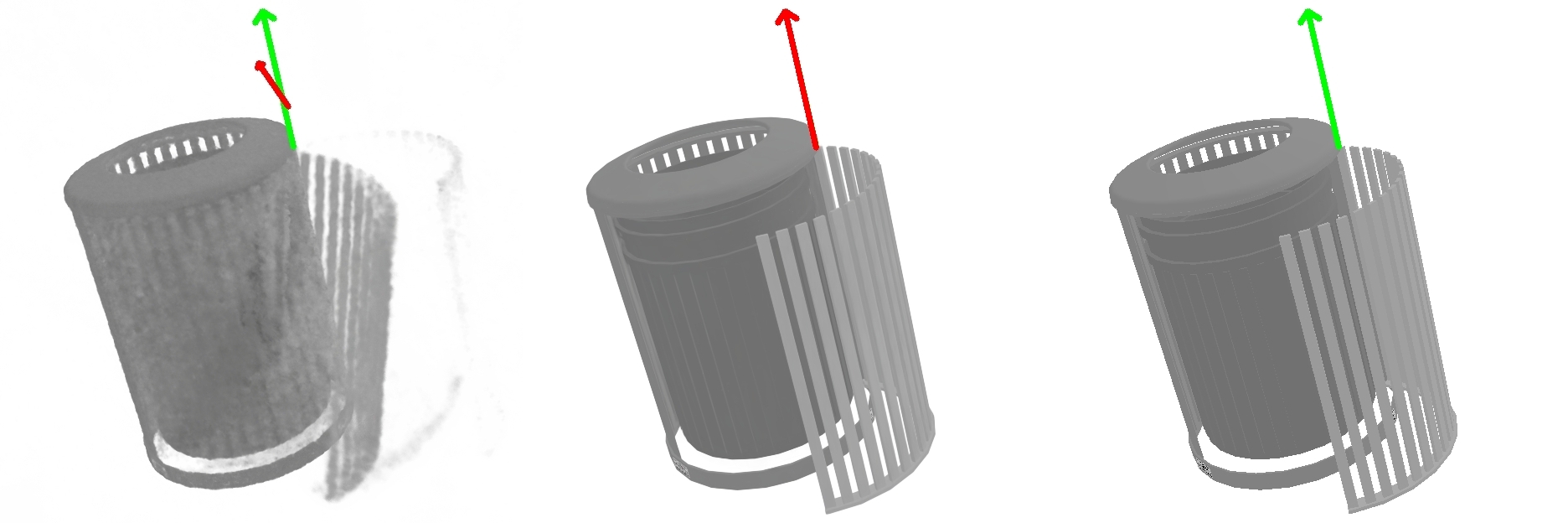} & \includegraphics[width=0.35\textwidth, align=c]{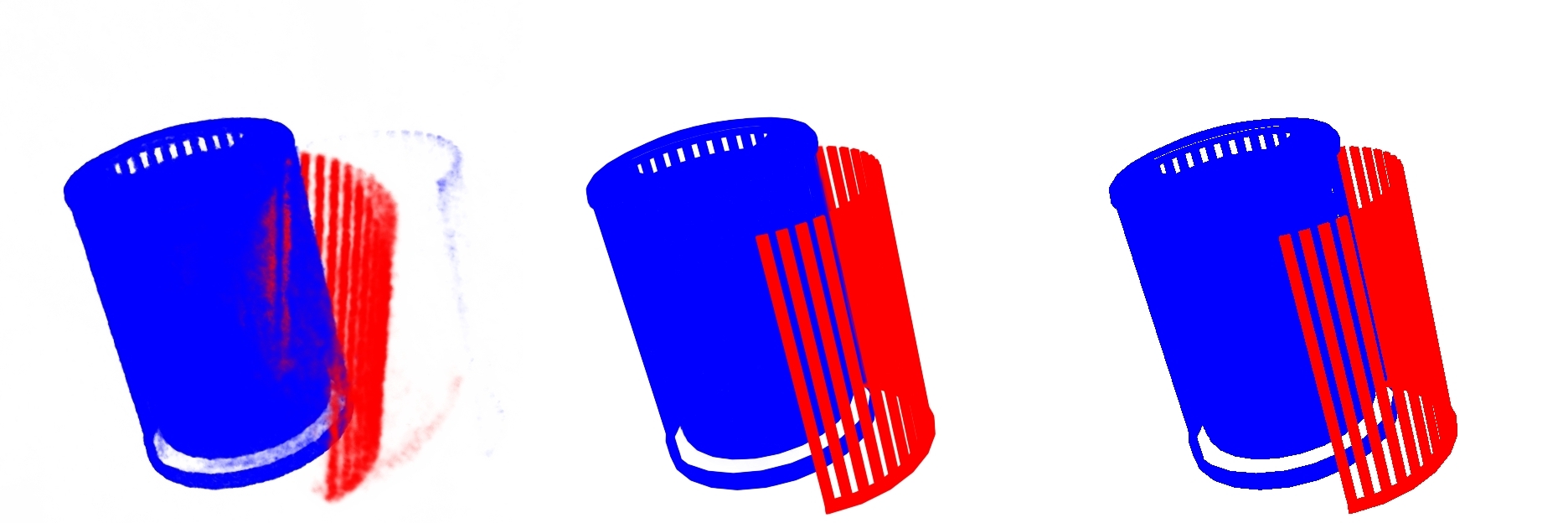} \\
        \bottomrule
    \end{tabular}
    \caption{Qualitative comparison on part-level reconstruction and articulation estimation.}
    \label{tab:qualitative-comparison}
\end{table*}

\subsection{Evaluation Metrics}
\paragraph{Articulation Estimation.}
We evaluate the accuracy of articulation estimation depending on its type.
\begin{itemize}
    \item A revolute articulation describes a rotation around some line in 3D space, parameterized by the pivot point $\mathbf{p}$ on the line, the line's directional axis $\mathbf{a}$, and the angle of rotation $\theta$. We report the angular error $\textrm{err}_\mathbf{a}$ (in $\times10^{-2}$ DEG) between the predicted and ground-truth axes, the geodesic distance $\textrm{err}_r$ (in $\times10^{-2}$ DEG) between the predicted and ground-truth rotations induced by the axis-angle ($\mathbf{a}$-$\theta$) pair, and the pivot point error $\textrm{err}_p$. Since the pivot can move arbitrarily along the axis, $\textrm{err}_p$ is computed as the closest distance between the predicted and ground-truth lines induced by the axis-pivot ($\mathbf{a}$-$\mathbf{p}$) pair.
    \item A prismatic articulation describes a translation along a specific direction, parameterized by the axis $\mathbf{a}$ of the translation direction and the translation distance $d$. We report the axis error $\textrm{err}_\mathbf{a}$, as in the revolute case, and the translation error $\textrm{err}_d$, which is the distance between the predicted and ground-truth translations induced by the axis-distance pair.
\end{itemize}
\paragraph{Part-Level Reconstruction.}
The evaluation of part-level reconstruction accuracy is threefold: photometric rendering quality, geometric accuracy, and part segmentation accuracy. For all these aspects, we use novel view synthesis as the surrogate task. We perform volume rendering for each view in the test split, generating outputs that include the RGB image, depth map, and part segmentation map. Photometric rendering quality is evaluated by reporting image metrics PSNR. Geometric accuracy is assessed by reporting the mean absolute error of the depth map (i.e., Depth MAE). Part segmentation accuracy is evaluated using the intersection-over-union (IoU) ratio for three categories: static part (IoU\textsubscript{s}), mobile part (IoU\textsubscript{m}), and background (IoU\textsubscript{bg}). The mean IoU (mIoU) is then reported as the average IoU across these categories. In addition, we also evaluate geometric accuracy through mesh reconstruction. To extract a mesh from 3DGS, we:
\begin{enumerate*}[label=(\arabic*)]
    \item render depth images from uniformly sampled spherical viewpoints,
    \item fuse rendered depths into a TSDF representation~\cite{curless1996volumetric,newcombe2011kinectfusion}, and
    \item extract the mesh using the Marching Cubes algorithm~\cite{lorensen1998marching}.
\end{enumerate*}
For each reconstruction, we separately extract the meshes corresponding to the static, mobile, and whole parts. For evaluation, we follow the procedure used in previous works~\cite{liu2023paris,weng2024neural}: \num{10000} points are uniformly sampled from both the reconstructed and the ground-truth meshes, and the Chamfer distance is computed for each category---static ($\textrm{CD}_s$), mobile ($\textrm{CD}_m$), and whole ($\textrm{CD}_w$).

\begin{figure}[!tb]
    \centering
    \includegraphics[width=\linewidth]{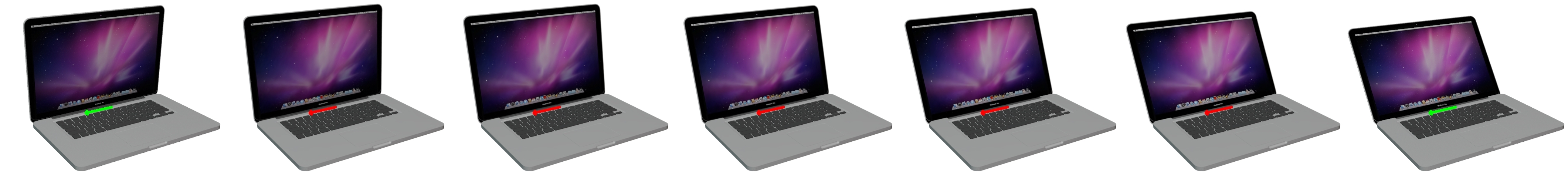} \\
    \includegraphics[width=\linewidth]{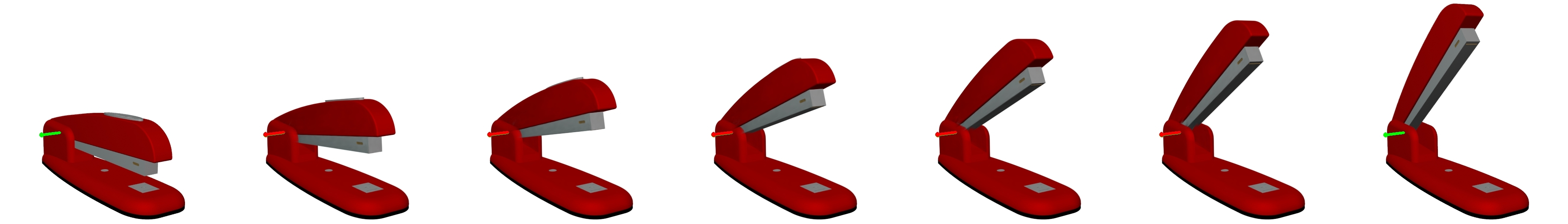} \\
    \includegraphics[width=\linewidth]{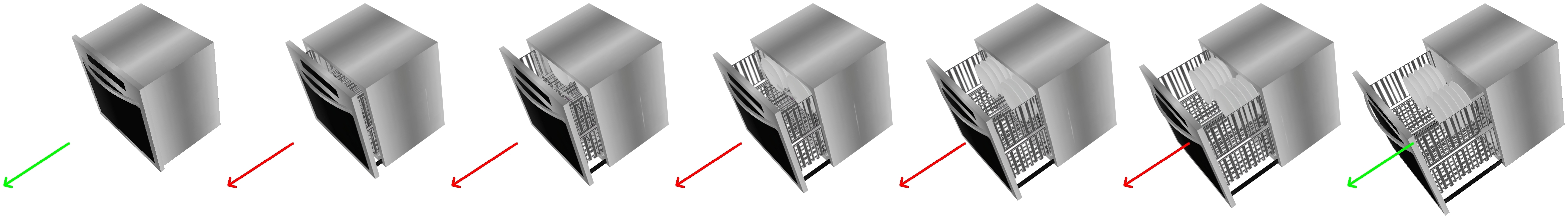} \\
    \caption{State interpolation from the estimated articulation.}
    \label{fig:qualitative-interpolation}
\end{figure}

\paragraph{Impact of Randomness.}
Randomness can play a significant role in the joint optimization of articulation and part reconstruction~\cite{liu2023paris,deng2024articulate}. In order to effectively demonstrate the robustness of a method, we conduct ten trials for each scene, each with a different fixed random seed. 
We report results averaged over the ten runs. Additionally, we notice that for a failed run converging to a local minimum, the evaluation metrics, such as $\textrm{err}_r$, are typically orders of magnitude worse than those of a successful run. To better reflect a method's performance, we report the number of successful runs out of the total number (i.e., ten), where a run is deemed successful if and only if each of the following criteria are met:%
\begin{enumerate*}[label=(\arabic*)]
    \item $\textrm{err}_\mathbf{a}<5$\textrm{ DEG};
    \item $\textrm{err}_\mathbf{p}<0.05$ for revolute articulation;
    \item $\textrm{err}_r<10\textrm{ DEG}$ for revolute articulation; and
    \item $\textrm{err}_t<0.05$ for prismatic articulation.
\end{enumerate*}
We then proceed to report the evaluation metrics detailed previously, \emph{averaging only over the successful runs}.

\subsection{Baselines}
\paragraph{PARIS~\cite{liu2023paris}} addresses the same challenge we tackle: reconstructing the part-level shape, appearance, and motion of an articulated object from multi-view RGB images captured at two articulation states. It employs a NeRF-based representation, modeling the object with static and mobile fields for each part, alongside a transformation to capture state transitions. These components are jointly optimized end-to-end, primarily through an image-rendering loss.

\paragraph{DTA~\cite{weng2024neural}} adopts a similar setup: reconstructing digital twins of articulated objects from multi-view RGB-D observations at two articulation states. It operates in two stages: Stage 1 optimizes neural object fields for each state using RGB-D images and extracts meshes; Stage 2 optimizes a part segmentation field and per-part motions using consistency, matching, and collision losses to determine point correspondences between states. \emph{\textbf{Note}: Unlike \alg, DTA requires ground-truth depth as input and produces plain object meshes unsuitable for photorealistic rendering.}

\subsection{Experiment Results}
\paragraph{Articulation Estimation.}  
We quantitatively evaluate \alg against baselines for articulation estimation accuracy, reporting scene-averaged results per articulation type on the PARIS-PMS dataset (\cref{tab:paris-pms-articulation-mean}) and the \alg-PMS dataset (\cref{tab:splart-pms-articulation}). \alg consistently outperforms the baselines across both datasets. Additionally, we qualitatively compare \alg with PARIS on selected scenes from these datasets in \cref{tab:qualitative-comparison}. In the ``color'' column, each image overlays the ground-truth articulation (green arrow) and the estimated articulation (red arrow), except in the ground-truth column. When the estimate aligns perfectly with the ground truth, only the red arrow appears; if significantly misaligned, only the green arrow is visible.
\paragraph{Part-Level Reconstruction.}  
We assess \alg’s part-level reconstruction accuracy against baselines, using novel view synthesis as a surrogate task. Scene-averaged results per articulation type are reported on the PARIS-PMS dataset (\cref{tab:paris-pms-synthesis-mean}) and the \alg-PMS dataset (\cref{tab:splart-pms-synthesis}). We also evaluate geometric accuracy through mesh reconstruction, presenting scene-averaged results per articulation type on the PARIS-PMS dataset (\cref{tab:paris-pms-articulation-mean}). Without depth supervision, which is required for the DTA baseline, \alg matches DTA’s performance, both substantially exceeding PARIS. Qualitative comparisons with PARIS, including RGB renderings and part segmentation maps, are provided for selected scenes from both datasets in \cref{tab:qualitative-comparison}. Further qualitative results on articulation synthesis appear in \cref{fig:qualitative-interpolation}.
\paragraph{Real-World Reconstructions.}  
We qualitatively evaluate \alg on a collected real-world dataset, with results shown in \cref{tab:qualitative-real}. Each image overlays the estimated articulation, visualized with a red arrow.
\paragraph{Ablation Studies.}
We perform extensive ablation studies to demonstrate the contribution of each stage. Specifically, we perform the following four ablations:
\begin{enumerate*}[(\arabic*)]
    \item No geometric supervision for mobility estimation, i.e. skipping Stage 2(a), tagged \alg-2a;
    \item No photometric supervision for mobility estimation, i.e., skipping Stage 2(b), tagged \alg-2b;
    \item No geometric supervision for articulation estimation, i.e. skipping Stage 3(a), tagged \alg-3a;
    \item No photometric supervision for articulation estimation, i.e., skipping Stage 3(b), tagged \alg-3b.
\end{enumerate*}
Like the main experiments, we conduct 10 runs for each scene, but only on the more challenging dataset of \alg-PMS. We report the articulation estimation results in \cref{tab:splart-pms-articulation}.

\begin{figure}[!t]
    \centering
    \includegraphics[width=\linewidth]{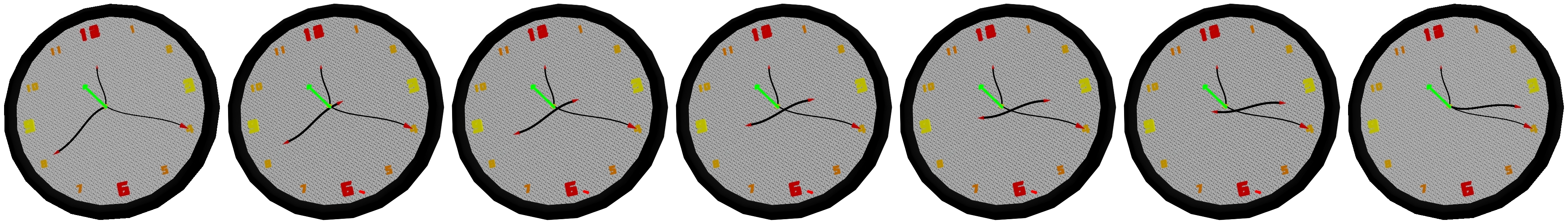} \\
    \includegraphics[width=\linewidth]{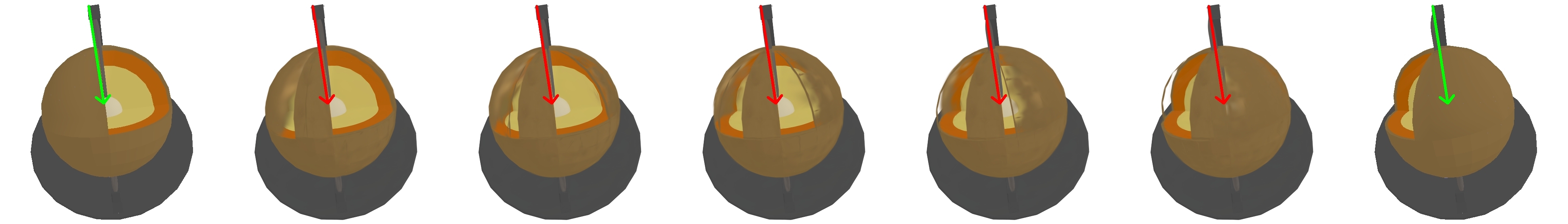} \\
    \caption{Failure cases due to inherent ambiguities in articulation.}
    \label{fig:qualitative-failure}
\end{figure}

\section{Conclusion}\label{sec:conclusion}
We present \alg, the first self-supervised method that utilizes 3D Gaussian Splatting for articulated object reconstruction from two-state RGB observations, without 3D supervision or category-specific priors. \alg delivers robust optimization and effectively handles challenging scenarios---outperforming prior approaches while eliminating the need for 3D supervision, articulation annotations, or semantic labels. Extensive evaluations on synthetic and real-world datasets show \alg's articulation estimation accuracy and view synthesis quality that surpass existing methods. Real-world experiments further validate its practicality, showcasing successful reconstructions of diverse articulated objects using only a handheld RGB camera. Our framework empowers non-expert users to effortlessly create high-fidelity, articulated digital twins, supporting applications in augmented/virtual reality and robotics. However, while \alg effectively addresses two-part articulations, it lacks direct support for multi-part articulated objects, though iteratively applying it to incremental state pairs may help. Some existing methods~\cite{weng2024neural, deng2024articulate, liu2025building} handle such objects, but require known part counts and single-level articulation structures. Future research will focus on a comprehensive adaptation to multi-level articulated objects. Additionally, inherent ambiguities may arise when multiple articulations explain state differences, as shown in Fig.~\ref{fig:qualitative-failure} with examples like a clock's hand rotating on an unexpected axis or a globe's surface spinning independently. Overall, \alg represents a scalable, robust, and practical solution that pushes the boundaries of articulated object reconstruction, laying a strong foundation for future work.

{
\small
\bibliographystyle{ieeenat_fullname}
\bibliography{references}
\flushcolsend
}

\clearpage
\ifforarxiv
    \appendix
\else
    \setcounter{page}{1}
    \maketitlesupplementary
\fi
\section{3D Gaussian Splatting}
3D Gaussian Splatting~\cite{kerbl20233dgs} (3DGS) is a method for reconstructing 3D scenes from posed images by representing the scene using Gaussian distributions in a continuous 3D space. Given a Gaussian blob parameterized by $(\mu,R,S,\sigma)$, where $\mu\in\mathbb{R}^3$ denotes the position of the center (mean) of the Gaussian,  $R\in\mathbb{R}^{3\times3}$ is a rotation matrix that denotes its orientation, $S\in\mathbb{R}^{3\times3}$ is the scale matrix (the scale matrix and rotation matrix determine the covariance matrix of the Gaussian), and $\sigma\in\mathbb{R}^+$ denotes its opacity. By design, the influence of a Gaussian on a point $\mathbf{x}$ is given by
\begin{multline}
    g(\mathbf{x}|\mu,R,S,\sigma)\\
    =\sigma\exp\left(-\frac{1}{2}(\mathbf{x}-\mu)^T\left(RSS^TR^T\right)^{-1}(\mathbf{x}-\mu)\right).
    \label{eq:1}
\end{multline}
In practice, 3D Gaussians are first projected to 2D given the camera view, while all Gaussians intersecting with a pixel's ray are sorted by depth for alpha-compositing. Please refer to \citet{kerbl20233dgs} and \citet{ye2024gsplatopensourcelibrarygaussian} for more technical details. Compared to Neural Radiance Fields~\cite{mildenhall2021nerf} (NeRFs), which represent a scene with radiance and density fields in the form of neural networks, 3DGS offers a much faster rendering speed (more than $100\times$ speed-up) thanks to the efficient rasterization of Gaussians.

\section{Articulating the Gaussians}

In \alg, we articulate a Gaussian blob by
\begin{enumerate*}[label=(\arabic*)]
    \item rotating by an angle $\theta$ around a line specified by $(\mathbf{p},\mathbf{a})$, where $\mathbf{p}$ denotes the pivot point and $\mathbf{a}$ denotes the axis direction; and
    \item translating along the same line by distance $d$.
\end{enumerate*}
We then need to update the parameters of the Gaussian $(\mu',R',S',\sigma')$ to reflect this articulation. Further, letting $\mathbf{d}$ be the camera-to-Gaussian direction computed after the articulated motion, we want to find out the actual direction $\mathbf{d}'$ with which to query the radiance field (i.e., a spherical function represented by Spherical Harmonics).

\begin{figure}[!t]
    \centering
    \includegraphics[width=\linewidth]{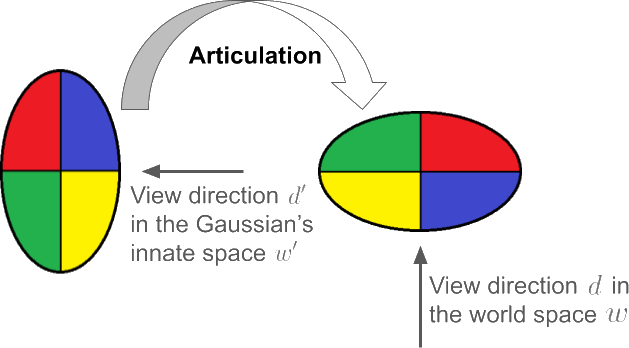}
    \caption{Since the radiance field is not modified by the articulation, we update the view direction when querying the Gaussian's radiance field in order to compensate for the articulation.}%
    \label{fig:articulated-gaussian}
\end{figure}

Let $\mathbf{x}'$ be the new point that results from applying the articulated motion to $\mathbf{x}$, which follows as
\begin{equation}
    \mathbf{x}'=R_{\mathbf{a},\theta}(\mathbf{x}-\mathbf{p})+\mathbf{p}+d\mathbf{a},
    \label{eq:2}
\end{equation}
where $R_{\mathbf{a},\theta}$ is the matrix form of the axis-angle rotation $(\mathbf{a},\theta)$.
By definition,
\begin{equation}
    g(\mathbf{x}' \mid \mu',R',S',\sigma') = g(\mathbf{x} \mid \mu,R,S,\sigma)\, \forall\mathbf{x}.
    \label{eq:3}
\end{equation}
By plugging Equations~\ref{eq:1} and \ref{eq:2} into Equation~\ref{eq:3}, the updated parameters follow as
\begin{subequations}
    \begin{align}
        \mu'&=R_{\mathbf{a},\theta}(\mu-\mathbf{p})+\mathbf{p}+d\mathbf{a},\\
        R'&=R_{\mathbf{a},\theta}R,\\
        S'&=S,\\
        \sigma'&=\sigma.
    \end{align}
\end{subequations}
Note that only the Gaussians are articulated, but not the radiance fields (since the coefficients of Spherical Harmonics are \emph{not} changed). To determine the actual direction $\mathbf{d}'$ with which to query the radiance field, imagine there is a world space $w'$ that follows the same articulated motion, meaning that $w'$ is stationary relative to the Gaussian blob. By definition, an articulated point $\mathbf{x}'$ given by Equation~\ref{eq:2} in the original world space $w$ is still denoted as point $\mathbf{x}$ in $w'$. So
\begin{equation}
    ^wT_{w'}(\mathbf{x})=R_{\mathbf{a},\theta}(\mathbf{x}-\mathbf{p})+\mathbf{p}+d\mathbf{a},
\end{equation}
where $^wT_{w'}(\cdot)$ denotes the transformation from $w'$ to $w$. Now, given view direction $\mathbf{d}$ in $w$, the actual query direction is represented via the Gaussian's innate space, which can be computed as
\begin{subequations}
   \begin{align}
    \mathbf{d}'&={}^{w'}T_w(\mathbf{d})\\
    &={R_{\mathbf{a},\theta}}^{-1}\mathbf{d}.
\end{align}
\end{subequations}
See Figure~\ref{fig:articulated-gaussian} for an illustration.

\section{Implementation}  
\alg leverages \emph{nerfstudio}~\cite{nerfstudio} and \emph{gsplat}~\cite{ye2024gsplatopensourcelibrarygaussian}, widely used open-source libraries for neural rendering and Gaussian splatting, respectively. We apply consistent hyper-parameters across all experiments, spanning synthetic and real-world datasets. Optimization stages vary in termination criteria: those driven by geometric consistency---Stages 2(a), 3(a), and 3(c)---halt upon convergence, assessed by the loss function’s rate of change, whereas those guided by photometric loss---Stages 1, 2(b), and 3(b)---stop after fixed iterations of \num{10000}, \num{5000}, and \num{10000}, respectively. By contrast, \emph{splatfacto} (\emph{nerfstudio}’s default 3DGS implementation) uses \num{30000} iterations. In Stage 3(a), we set $K^\textrm{m} = K^\textrm{cm} = 3$ by default, with each optimization trial lasting 10 seconds to 1 minute on an RTX 2080 Ti. Consequently, \alg’s total training time, approximately 20--30 minutes, is roughly twice that of \emph{splatfacto}. For comparison, PARIS~\cite{liu2023paris} requires 15--20 minutes, while DTA~\cite{weng2024neural}, including its LoFTR pixel-matching step, takes 35--45 minutes. At inference, rendering for novel view and articulation synthesis achieves 60--100 frames per second, varying with scene complexity due to the increased Gaussian count for intricate objects.
\section{Robustness of Geometric Consistency}%
\begin{figure*}[!t]
    \centering
    \includegraphics[width=\linewidth]{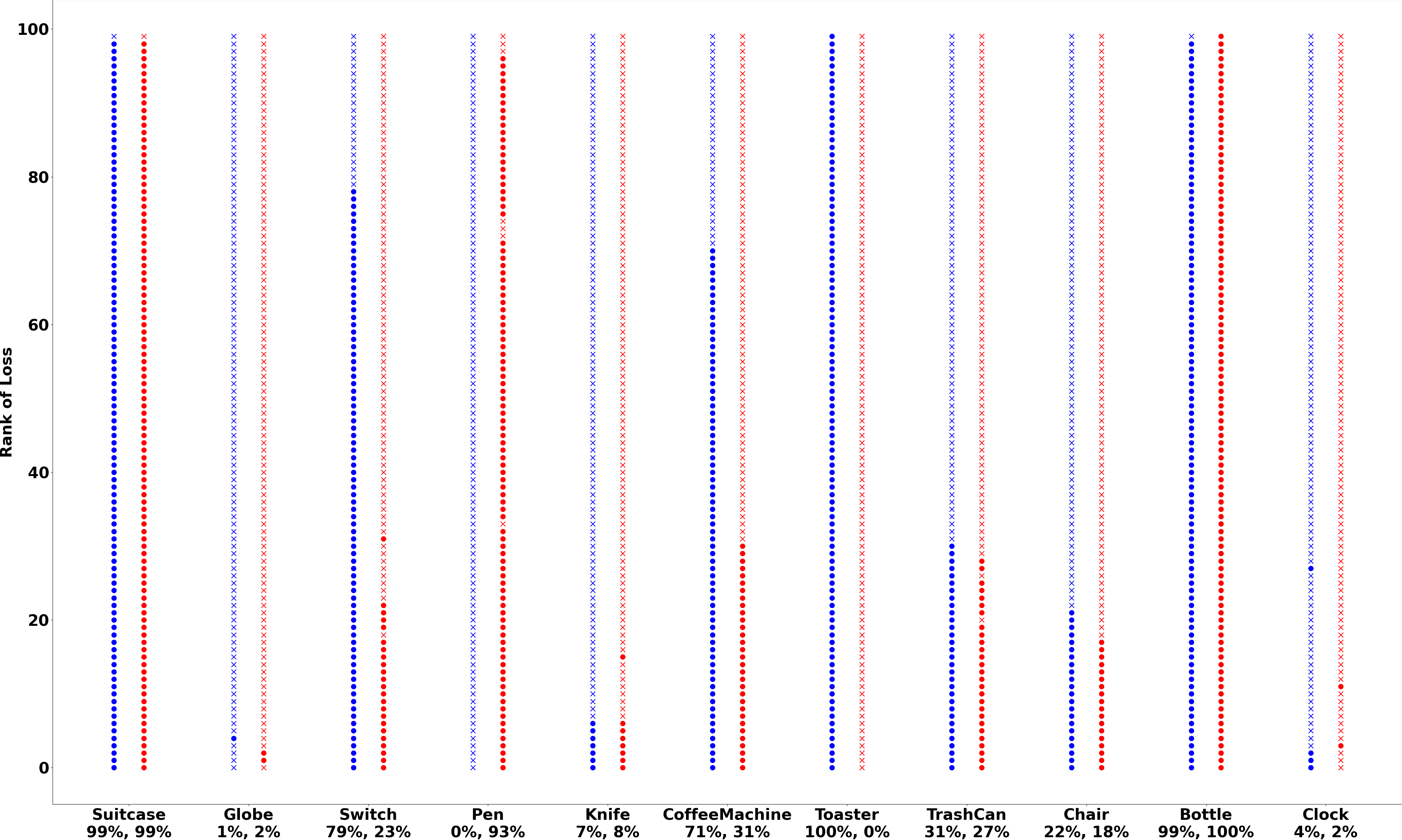}
    \caption{An analysis of the robustness of geometric consistency for articulation estimation. The horizontal axis represents different scenes, each with two columns for the two optimization formulations, where \textcolor{blue}{blue} is used for optimization trials with mobile-only geometric consistency, and \textcolor{red}{red} for those with cross-mobile geometric consistency. The vertical axis represents the loss rank (lower is better). For each optimization trial, $\bullet$ denotes success, and $\times$ denotes failure. For each scene, the average success rates for both optimization formulations are shown under the scene name. Note that these success rates are the averages over the independent optimization trials and should not be confused with that of Stage 3(a) as a whole. We observe that successful trials $\bullet$ are often associated with lower losses than are failed trials $\times$, and that both optimization formulations are necessary to ensure overall robustness (e.g., \emph{pen} requires \textcolor{red}{cross-mobile geometric consistency}, while \emph{toaster} requires \textcolor{blue}{mobile-only geometric consistency}).}
    \label{fig:optim-trials}
\end{figure*}
In Stage 3(a), we propose a practical strategy that involves multiple attempts using both mobile-only and cross-mobile geometric consistency for robust articulation estimation. For this approach to be effective, two prerequisites must be met:
\begin{enumerate*}[label=(\arabic*)]
    \item a correct articulation should induce a lower loss than most, if not all, incorrect articulation estimates;  and
    \item the optimization should have a high likelihood of converging to the global optimum (indicated by the correctness of the articulation) for each randomized attempt, ensuring that a small number of trials is sufficient and robust.
\end{enumerate*}
To verify that both prerequisites are satisfied, we design a dedicated experiment that involves 200 independent optimization trials: 100 using mobile-only geometric consistency and 100 using cross-mobile geometric consistency, both performed based on the model checkpoint saved at the end of Stage 2(b). A trial is deemed successful if it meets a similar but slightly relaxed criterion (since the estimated articulation is still coarse at this stage). Additionally, we rank the loss of each trial in comparison to the other trials performed under the same formulation. In light of the prerequisites above, we are interested in seeing whether successful trials tend to exhibit lower loss compared to those that are unsuccessful. Figure~\ref{fig:optim-trials} visualizes these results as a scatter plot for a set of objects from the \alg-PMS dataset.

\section{Additional Qualitative Results}
\begin{figure*}
    \centering
    \includegraphics[width=\linewidth]{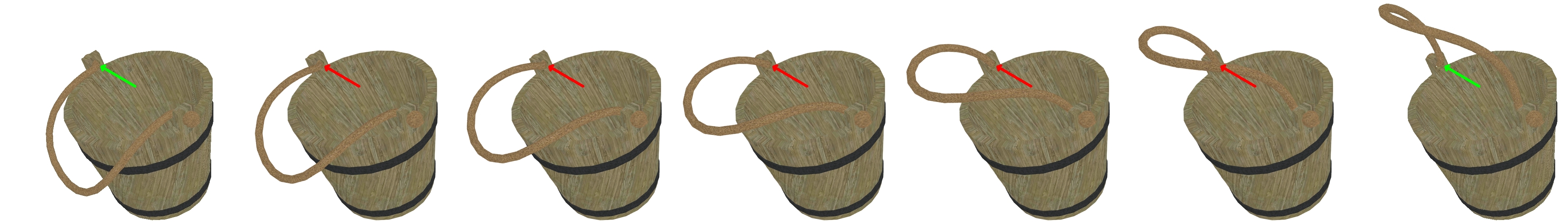} \\
    \includegraphics[width=\linewidth]{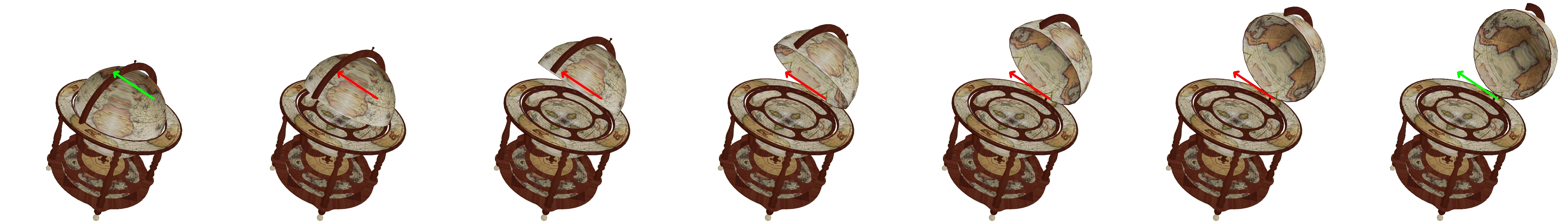} \\
    \includegraphics[width=\linewidth]{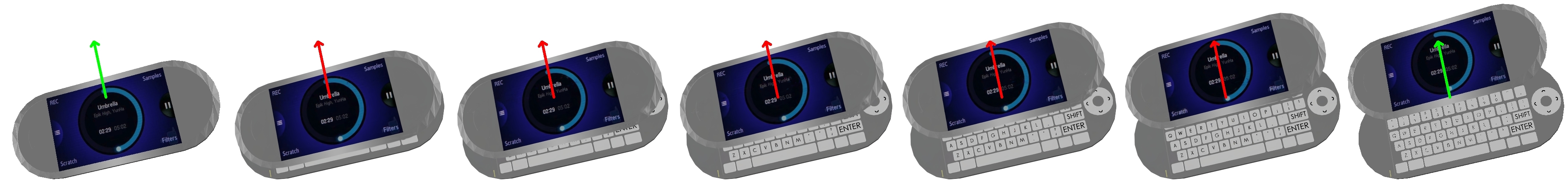}\\
    \quad\qquad State 0 \hfill State 1 \quad\qquad 
    \caption{Additional visualizations of state interpolation for three objects along with their estimated articulation models. The left- and right-most images correspond to the ground-truth of the two end articulation states, annotated with the ground-truth articulation using \textcolor{teal}{green arrows}. The intermediate images are rendered results obtained by interpolating the articulation state, annotated with both the ground-truth articulation (\textcolor{teal}{green arrows}) and the estimated articulation (\textcolor{red}{red arrows}). Due to the high accuracy of the articulation estimation, the ground-truth \textcolor{teal}{green arrows} are hidden by the estimated \textcolor{red}{red arrows} in the interpolated images.}
    \label{fig:qualitative-interpolation-more}
\end{figure*}

\begin{figure*}[!ht]
    \centering
    \begin{tabular}{ccc}
        \toprule
        \multirow{2}{*}{Input States} & Color & Part Segmentation \\
        \cmidrule(lr){2-3}
        & \hspace{0.5em} PARIS \hspace{2em} \alg \hspace{1em} Ground-truth & \hspace{0.5em} PARIS \hspace{2em} \alg \hspace{1em} Ground-truth \\
        \midrule
        \multicolumn{1}{c|}{\includegraphics[width=0.22\textwidth, align=c]{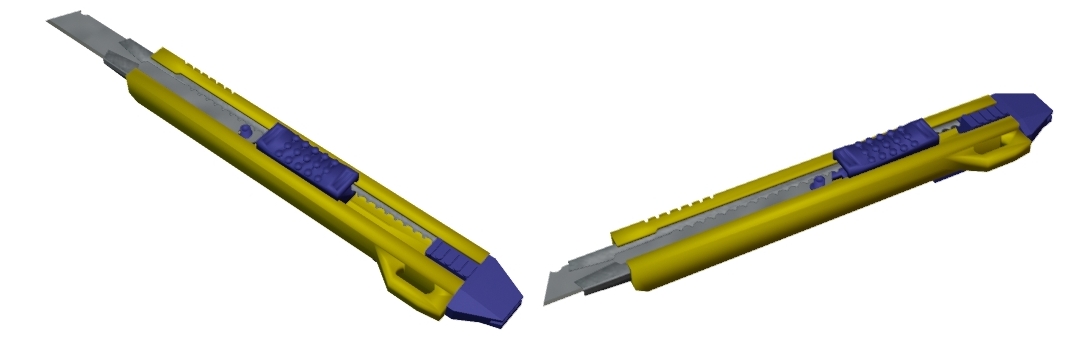}} & \includegraphics[width=0.35\textwidth, align=c]{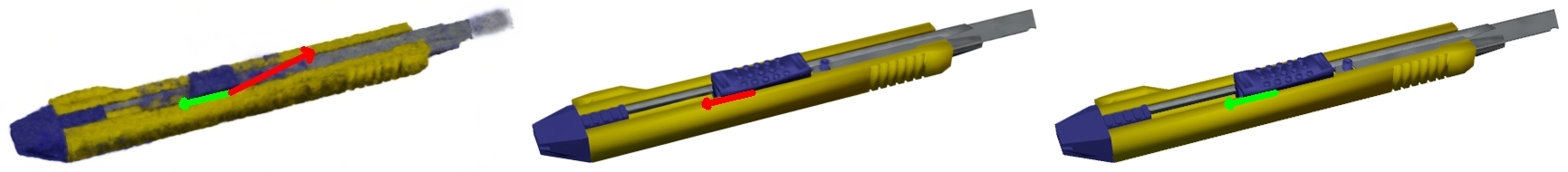} & \includegraphics[width=0.35\textwidth, align=c]{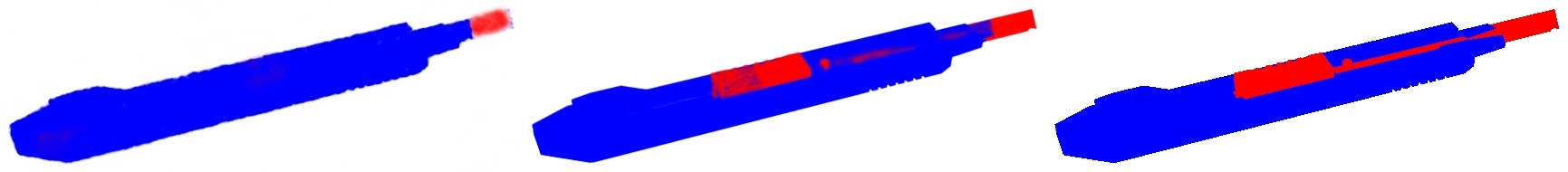} \\
        \multicolumn{1}{c|}{\includegraphics[width=0.22\textwidth, align=c]{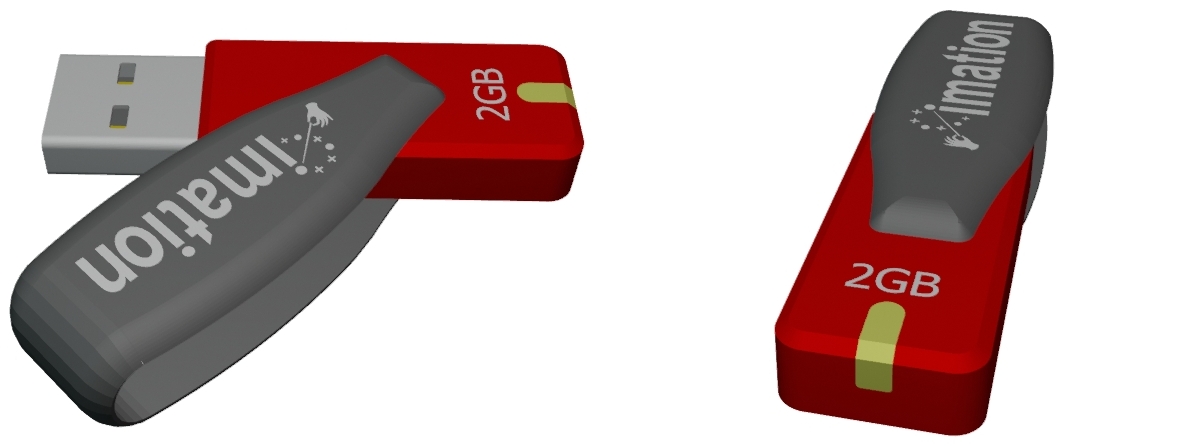}} & \includegraphics[width=0.35\textwidth, align=c]{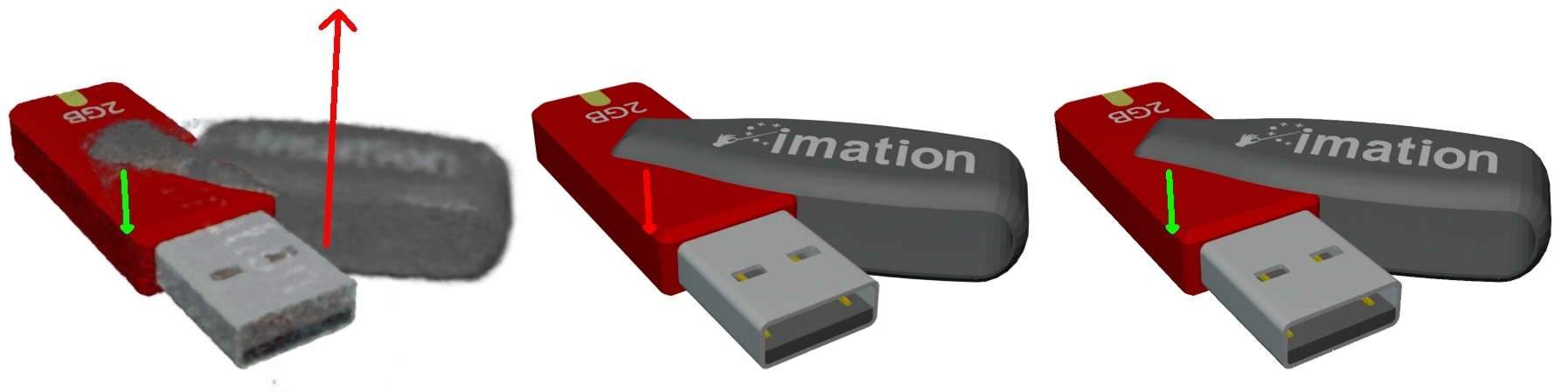} & \includegraphics[width=0.35\textwidth, align=c]{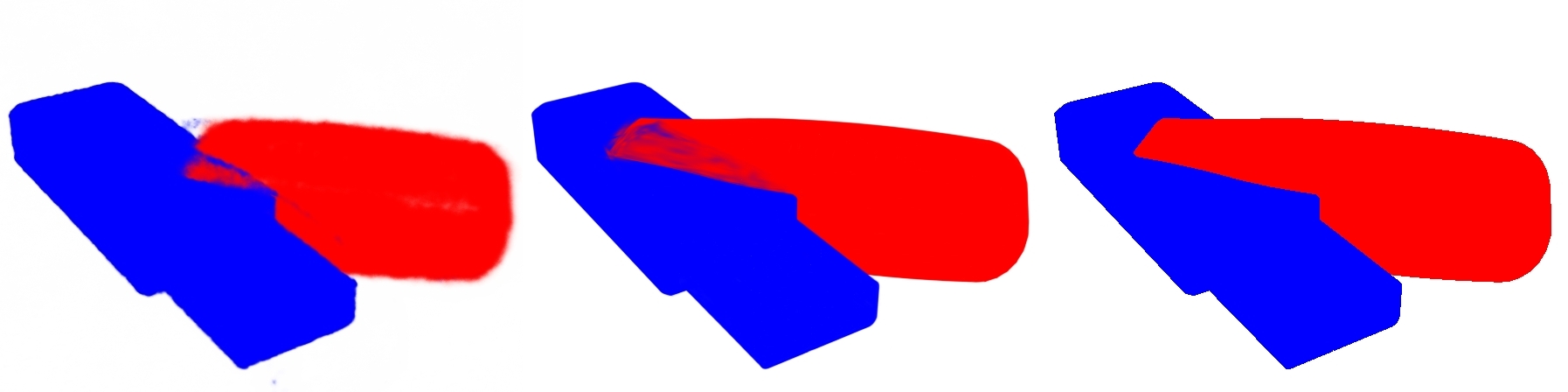} \\
        \multicolumn{1}{c|}{\includegraphics[width=0.22\textwidth, align=c]{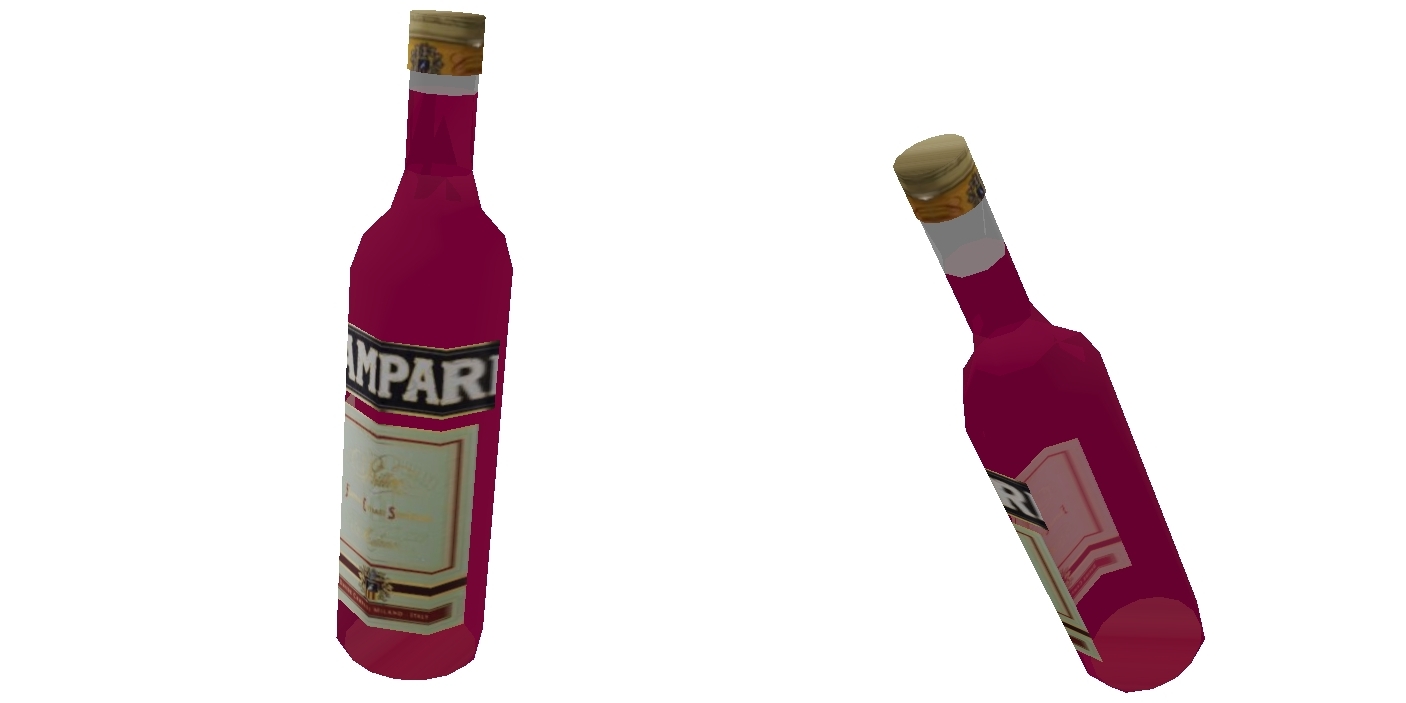}} & \includegraphics[width=0.35\textwidth, align=c]{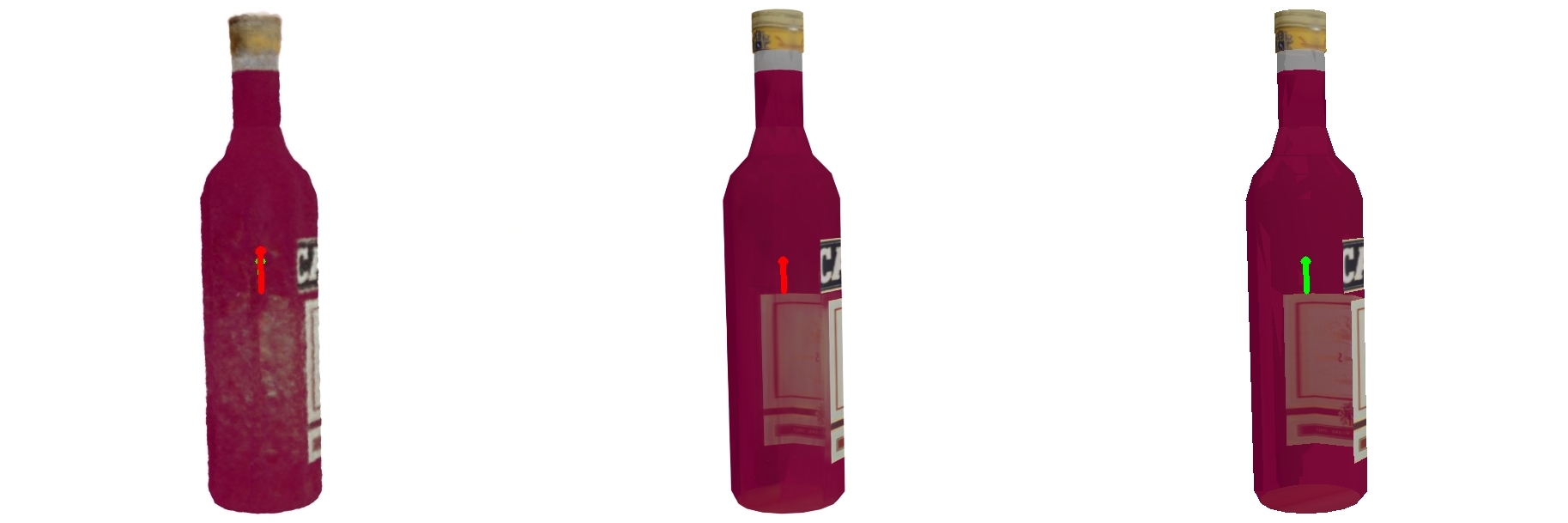} & \includegraphics[width=0.35\textwidth, align=c]{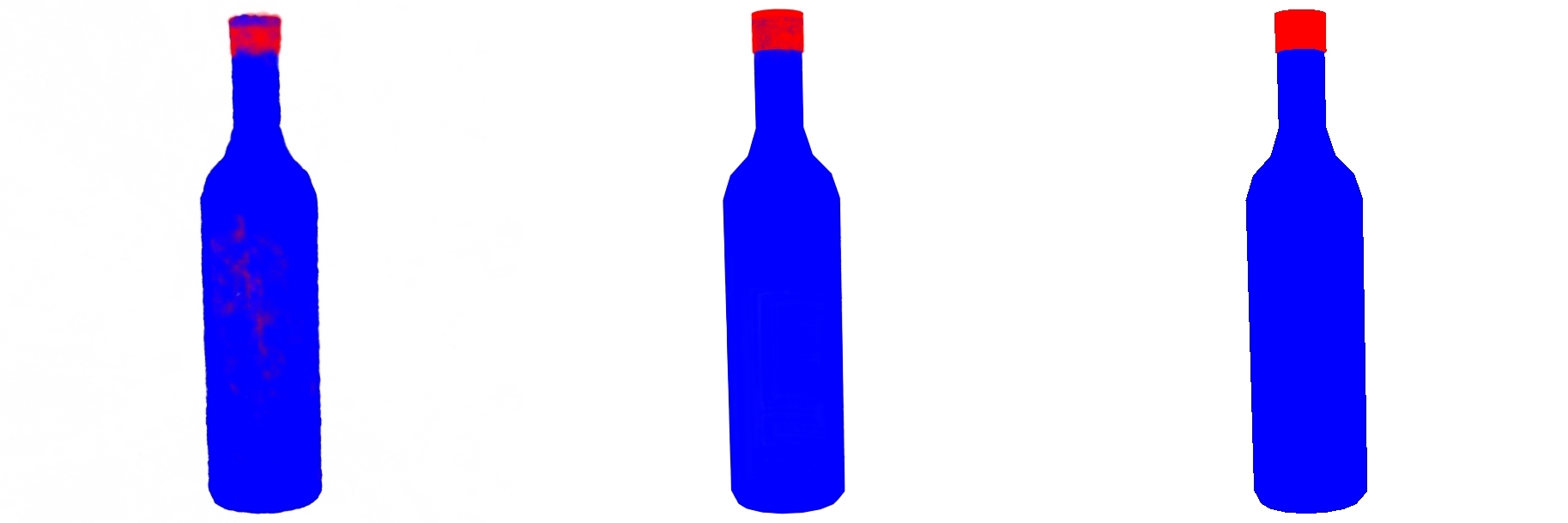} \\
        
        \multicolumn{1}{c|}{\includegraphics[width=0.22\textwidth, align=c]{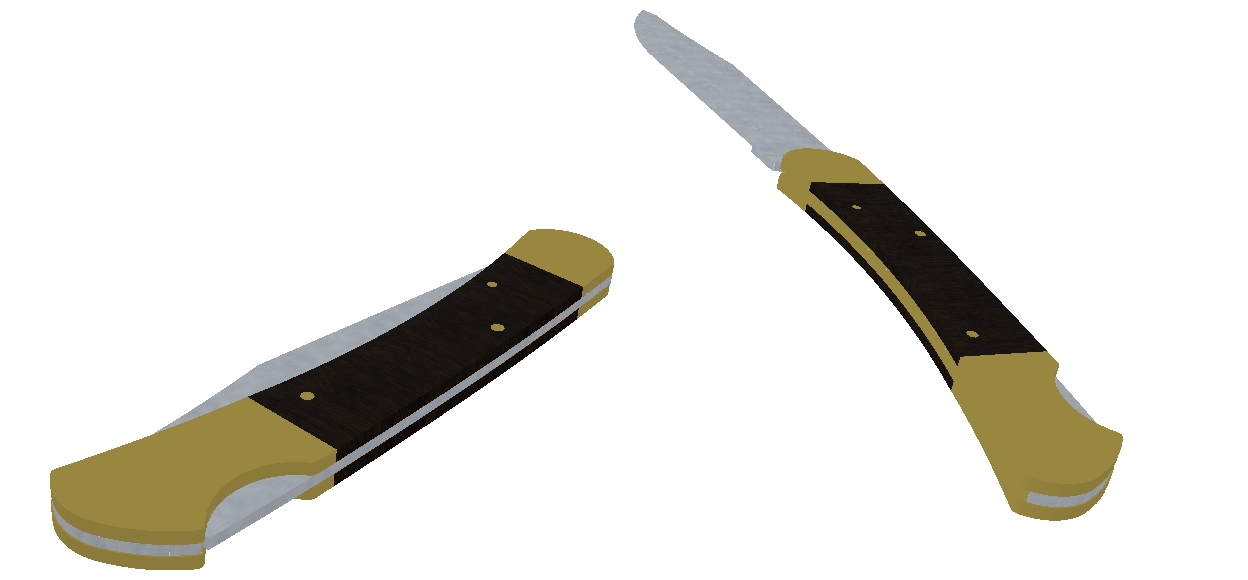}} & \includegraphics[width=0.35\textwidth, align=c]{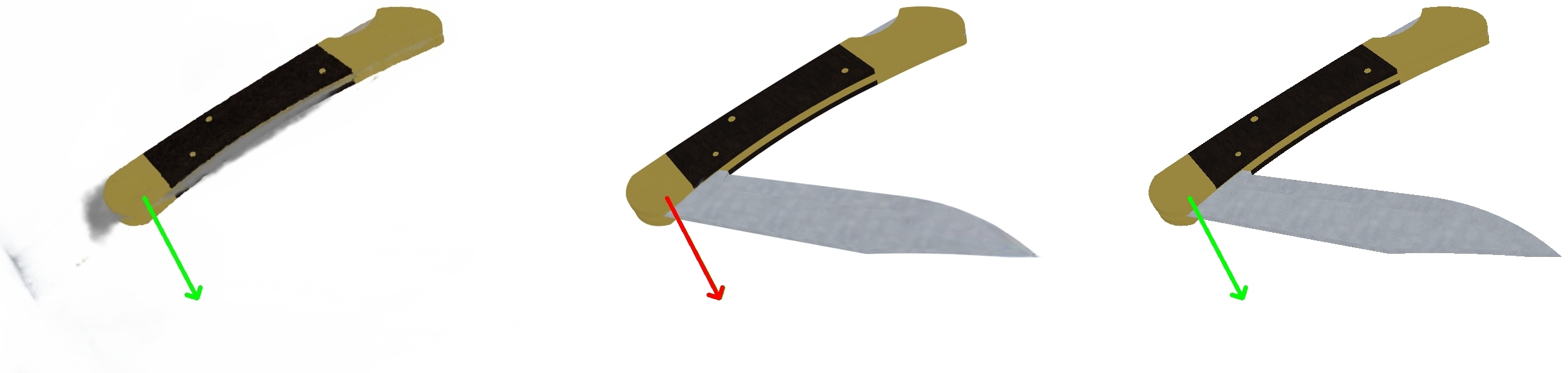} & \includegraphics[width=0.35\textwidth, align=c]{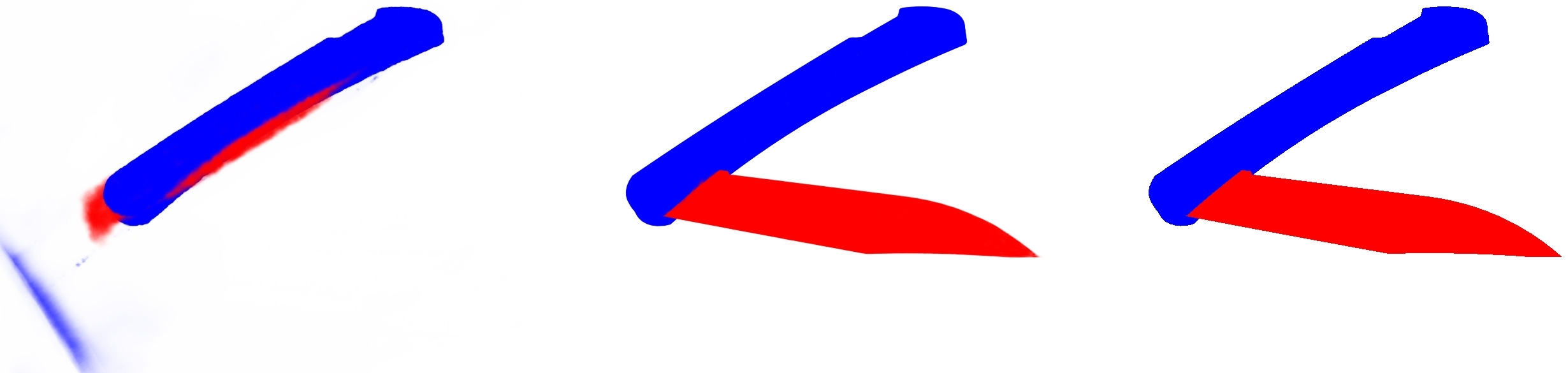} \\
        \bottomrule
    \end{tabular}
    \caption{Additional qualitative comparisons of part-level reconstruction and articulation estimation.}
    \label{tab:qualitative-comparison-more}
\end{figure*}

We present more qualitative results, including comparison with PARIS~\cite{liu2023paris} on part-level reconstruction and articulation estimation in Figure~\ref{tab:qualitative-comparison-more}, and view synthesis for interpolated articulation states in Figure~\ref{fig:qualitative-interpolation-more}.

\section{Per-Scene Quantitative Results}
We report quantitative comparison of \alg and PARIS for each scene in both PARIS-PMS and \alg-PMS. Specifically, see \cref{tab:paris-pms-articulation}, \cref{tab:splart-pms-articulation-revolute}, and \cref{tab:splart-pms-articulation-prismatic} for articulation estimation results on PARIS-PMS, revolute cases of \alg-PMS, and prismatic cases of \alg-PMS, respectively. For novel view and articulation synthesis, we further include SSIM~\cite{wang2004image} and LPIPS~\cite{zhang2018unreasonable} metrics (in addition to PSNR~\cite{jain1989fundamentals}) for the evaluation of rendering quality, and per-category intersection-over-union ratios (in addition to mIoU): IoU\textsubscript{s} for the static part, IoU\textsubscript{m} for the mobile part, and IoU\textsubscript{bg} for the background. See \cref{tab:paris-pms-synthesis}, \cref{tab:splart-pms-synthesis-revolute}, and \cref{tab:splart-pms-synthesis-prismatic} for results on PARIS-PMS, revolute cases of \alg-PMS, and prismatic cases of \alg-PMS, respectively. 

\begin{table*}[ht]
    \resizebox{\linewidth}{!}{
        \begin{tabular}{cllcS[table-format=3.2(3.2)]S[table-format=1.2(1.2)]S[table-format=3.2(2.2)]S[table-format=1.2(1.2)]S[table-format=2.2(2.2)]S[table-format=3.2(2.2)]S[table-format=2.2(1.2)]}
            \toprule
            Type & Scene & Method & \makecell{Success \\ Rate} $\big\uparrow$ & {\makecell{$\textrm{err}_a$ \\ ($\times10^{-2}$ DEG)} $\big\downarrow$} & {\makecell{$\textrm{err}_p$ \\ $(\times10^{-3})$} $\big\downarrow$} & {\makecell{$\textrm{err}_r$ \\ ($\times10^{-2}$ DEG)} $\big\downarrow$} & {\makecell{$\textrm{err}_t$ \\ $(\times10^{-3})$} $\big\downarrow$} & {\makecell{$\textrm{CD}_s$ \\ $(\times10^{-3})$} $\big\downarrow$} & {\makecell{$\textrm{CD}_m$ \\ $(\times10^{-3})$} $\big\downarrow$} & {\makecell{$\textrm{CD}_w$ \\ $(\times10^{-3})$} $\big\downarrow$} \\
            \midrule
            \multirow{27}{*}{\rotatebox{90}{Revolute}} & \multirow{3}{*}{USB} & PARIS & $0/10$ & {F} & {F} & {F} & {N/A} & {F} & {F} & {F} \\
            & & DTA$^\dagger$ & $10/10$ & 9.30 (1.29) & 0.39 (0.31) & 14.81 (3.49) & {N/A} & 2.62 (0.06) & 1.48 (0.03) & 1.36 (0.02) \\
            & & SplArt & $10/10$ & 4.74 (0.32) & 0.26 (0.05) & 3.99 (0.26) & {N/A} & 0.89 (0.08) & 0.83 (0.11) & 0.91 (0.03) \\
            \dashrule{2-11}
            & \multirow{3}{*}{foldchair} & PARIS & $3/10$ & 103.68 (15.95) & 4.69 (4.77) & 195.70 (67.55) & {N/A} & 0.45 (0.09) & 45.34 (4.62) & 14.85 (1.00) \\
            & & DTA$^\dagger$ & $10/10$ & 3.34 (1.43) & 0.50 (0.31) & 9.72 (3.61) & {N/A} & 0.18 (0.00) & 0.14 (0.00) & 0.26 (0.00) \\
            & & SplArt & $10/10$ & 5.47 (0.22) & 0.12 (0.03) & 8.93 (0.34) & {N/A} & 1.53 (0.68) & 0.33 (0.01) & 0.31 (0.00) \\
            \dashrule{2-11}
            & \multirow{3}{*}{fridge} & PARIS & $9/10$ & 105.51 (50.87) & 6.04 (4.17) & 121.25 (45.78) & {N/A} & 2.97 (0.19) & 39.11 (2.07) & 11.62 (0.54) \\
            & & DTA$^\dagger$ & $10/10$ & 6.67 (2.79) & 0.51 (0.32) & 12.07 (2.08) & {N/A} & 0.63 (0.01) & 0.29 (0.01) & 0.70 (0.00) \\
            & & SplArt & $10/10$ & 4.14 (0.56) & 0.06 (0.05) & 5.81 (0.34) & {N/A} & 2.23 (0.05) & 1.21 (0.03) & 1.95 (0.04) \\
            \dashrule{2-11}
            & \multirow{3}{*}{laptop} & PARIS & $10/10$ & 117.26 (94.05) & 6.89 (5.45) & 99.11 (51.26) & {N/A} & 0.61 (0.33) & 32.43 (2.19) & 15.08 (1.41) \\
            & & DTA$^\dagger$ & $10/10$ & 6.65 (1.73) & 1.33 (0.52) & 11.81 (3.62) & {N/A} & 0.31 (0.00) & 0.14 (0.00) & 0.34 (0.00) \\
            & & SplArt & $10/10$ & 3.20 (0.55) & 0.59 (0.05) & 4.89 (0.28) & {N/A} & 0.24 (0.00) & 0.34 (0.01) & 0.35 (0.01) \\
            \dashrule{2-11}
            & \multirow{3}{*}{oven} & PARIS & $10/10$ & 161.16 (112.94) & 3.44 (3.81) & 95.15 (55.90) & {N/A} & 9.89 (0.93) & 156.89 (43.24) & 10.90 (1.70) \\
            & & DTA$^\dagger$ & $10/10$ & 19.35 (4.10) & 1.57 (0.60) & 11.03 (2.19) & {N/A} & 4.61 (0.07) & 0.44 (0.01) & 4.26 (0.06) \\
            & & SplArt & $10/10$ & 1.45 (0.32) & 0.90 (0.06) & 2.95 (0.27) & {N/A} & 7.10 (0.44) & 1.89 (3.52) & 6.02 (0.32) \\
            \dashrule{2-11}
            & \multirow{3}{*}{scissor} & PARIS & $0/10$ & {F} & {F} & {F} & {N/A} & {F} & {F} & {F} \\
            & & DTA$^\dagger$ & $10/10$ & 7.28 (5.05) & 4.84 (7.66) & 56.56 (95.23) & {N/A} & 1.67 (3.41) & 5.06 (1399) & 0.42 (0.00) \\
            & & SplArt & $10/10$ & 2.24 (0.17) & 0.24 (0.06) & 2.18 (0.26) & {N/A} & 0.37 (0.01) & 0.20 (0.01) & 0.25 (0.00) \\
            \dashrule{2-11}
            & \multirow{3}{*}{stapler} & PARIS & $0/10$ & {F} & {F} & {F} & {N/A} & {F} & {F} & {F} \\
            & & DTA$^\dagger$ & $9/10$ & 7.06 (5.68) & 1.81 (1.16) & 11.76 (11.74) & {N/A} & 2.93 (0.24) & 2.16 (0.89) & 2.03 (0.03) \\
            & & SplArt & $10/10$ & 4.64 (0.43) & 0.75 (0.08) & 4.09 (0.28) & {N/A} & 1.17 (0.02) & 2.12 (0.14) & 1.05 (0.02) \\
            \dashrule{2-11}
            & \multirow{3}{*}{washer} & PARIS & $0/10$ & {F} & {F} & {F} & {N/A} & {F} & {F} & {F} \\
            & & DTA$^\dagger$ & $10/10$ & 38.95 (11.24) & 3.91 (2.63) & 27.51 (8.30) & {N/A} & 4.68 (0.11) & 0.40 (0.01) & 4.46 (0.12) \\
            & & SplArt & $10/10$ & 3.76 (0.48) & 0.24 (0.15) & 5.95 (0.67) & {N/A} & 19.14 (1.27) & 1.59 (2.40) & 17.92 (1.22) \\
            \dashrule{2-11}
            & \multirow{3}{*}{mean} & PARIS & $40.0\%$ & 121.90 (68.45) & 5.27 (4.55) & 127.80 (55.12) & {N/A} & 3.48 (0.39) & 68.44 (13.03) & 13.11 (1.16) \\
            & & DTA$^\dagger$ & $98.8\%$ & 12.32 (4.16) & 1.86 (1.69) & 19.41 (16.28) & {N/A} & 2.20 (0.49) & 1.27 (1.87) & 1.73 (0.03) \\
            & & SplArt & $100.0\%$ & 3.70 (0.38) & 0.40 (0.07) & 4.85 (0.33) & {N/A} & 4.08 (0.32) & 1.06 (0.78) & 3.59 (0.21) \\
            \cmidrule(lr){1-11}
            \multirow{9}{*}{\rotatebox{90}{Prismatic}} & \multirow{3}{*}{blade} & PARIS & $0/10$ & {F} & {N/A} & {N/A} & {F} & {F} & {F} & {F} \\
            & & DTA$^\dagger$ & $10/10$ & 25.72 (4.58) & {N/A} & {N/A} & 0.94 (0.14) & 0.49 (0.01) & 31.11 (0.55) & 0.37 (0.01) \\
            & & SplArt & $9/10$ & 2.04 (0.61) & {N/A} & {N/A} & 0.22 (0.04) & 0.44 (0.01) & 26.60 (0.85) & 0.44 (0.01) \\
            \dashrule{2-11}
            & \multirow{3}{*}{storage} & PARIS & $10/10$ & 27.97 (13.09) & {N/A} & {N/A} & 4.28 (3.02) & 9.21 (1.94) & 151.78 (35.00) & 7.99 (0.49) \\
            & & DTA$^\dagger$ & $10/10$ & 6.79 (2.50) & {N/A} & {N/A} & 1.37 (0.13) & 4.88 (0.07) & 0.36 (0.00) & 4.07 (0.06) \\
            & & SplArt & $10/10$ & 2.69 (0.26) & {N/A} & {N/A} & 0.45 (0.04) & 11.60 (0.38) & 6.94 (1.43) & 6.94 (0.27) \\
            \dashrule{2-11}
            & \multirow{3}{*}{mean} & PARIS & $50.0\%$ & 27.97 (13.09) & {N/A} & {N/A} & 4.28 (3.02) & 9.21 (1.94) & 151.78 (35.00) & 7.99 (0.49) \\
            & & DTA$^\dagger$ & $100.0\%$ & 16.26 (3.54) & {N/A} & {N/A} & 1.15 (0.14) & 2.69 (0.04) & 15.74 (0.28) & 2.22 (0.03) \\
            & & SplArt & $95.0\%$ & 2.36 (0.44) & {N/A} & {N/A} & 0.33 (0.04) & 6.02 (0.20) & 16.77 (1.14) & 3.69 (0.14) \\
            \cmidrule(lr){1-11}
            \multicolumn{2}{c}{\multirow{3}{*}{Overall}} & PARIS & $42.0\%$ & 103.12 (57.38) & 5.27 (4.55) & 127.80 (55.12) & 4.28 (3.02) & 4.63 (0.70) & 85.11 (17.42) & 12.09 (1.03) \\
            & & DTA$^\dagger$ & $99.0\%$ & 13.11 (4.04) & 1.86 (1.69) & 19.41 (16.28) & 1.15 (0.14) & 2.30 (0.40) & 4.16 (1.55) & 1.83 (0.03) \\
            & & SplArt & $99.0\%$ & 3.44 (0.39) & 0.40 (0.07) & 4.85 (0.33) & 0.33 (0.04) & 4.47 (0.30) & 4.20 (0.85) & 3.61 (0.19) \\
            \bottomrule
        \end{tabular}
    }
    \caption{PARIS-PMS Articulation and Mesh Reconstruction Metrics. $^\dagger$DTA requires ground-truth depth.}
    \label{tab:paris-pms-articulation}
\end{table*}

\begin{table*}[ht]
    \centering
    \begin{tabular}{cllcccccccc}
        \toprule
        Type & Scene & Method & PSNR $\uparrow$ & SSIM $\uparrow$ & LPIPS $\downarrow$ & Depth MAE $\downarrow$ & IoU\textsubscript{s} $\uparrow$ & IoU\textsubscript{m} $\uparrow$ & IoU\textsubscript{bg} $\uparrow$ & mIoU $\uparrow$ \\
        \midrule
        \multirow{27}{*}{\rotatebox{90}{Revolute}} & \multirow{3}{*}{USB} & PARIS & F & F & F & F & F & F & F & F \\
        & & DTA$^\dagger$ & N/A & N/A & N/A & $0.022$ & $0.864$ & $0.892$ & $0.996$ & $0.917$ \\
        & & SplArt & $45.70$ & $0.996$ & $0.0029$ & $0.026$ & $0.982$ & $0.979$ & $1.000$ & $0.987$ \\
        \dashrule{2-11}
        & \multirow{3}{*}{foldchair} & PARIS & $33.48$ & $0.936$ & $0.0844$ & $0.096$ & $0.961$ & $0.939$ & $0.994$ & $0.965$ \\
        & & DTA$^\dagger$ & N/A & N/A & N/A & $0.041$ & $0.915$ & $0.950$ & $0.988$ & $0.951$ \\
        & & SplArt & $43.98$ & $0.992$ & $0.0119$ & $0.037$ & $0.981$ & $0.980$ & $0.999$ & $0.987$ \\
        \dashrule{2-11}
        & \multirow{3}{*}{fridge} & PARIS & $32.22$ & $0.967$ & $0.0766$ & $0.087$ & $0.973$ & $0.866$ & $0.997$ & $0.945$ \\
        & & DTA$^\dagger$ & N/A & N/A & N/A & $0.034$ & $0.957$ & $0.823$ & $0.992$ & $0.924$ \\
        & & SplArt & $41.40$ & $0.996$ & $0.0101$ & $0.045$ & $0.986$ & $0.917$ & $0.999$ & $0.967$ \\
        \dashrule{2-11}
        & \multirow{3}{*}{laptop} & PARIS & $31.64$ & $0.964$ & $0.0758$ & $0.057$ & $0.907$ & $0.959$ & $0.997$ & $0.954$ \\
        & & DTA$^\dagger$ & N/A & N/A & N/A & $0.014$ & $0.938$ & $0.964$ & $0.997$ & $0.966$ \\
        & & SplArt & $40.90$ & $0.995$ & $0.0096$ & $0.026$ & $0.931$ & $0.975$ & $0.999$ & $0.968$ \\
        \dashrule{2-11}
        & \multirow{3}{*}{oven} & PARIS & $31.48$ & $0.950$ & $0.1048$ & $0.132$ & $0.977$ & $0.891$ & $0.996$ & $0.955$ \\
        & & DTA$^\dagger$ & N/A & N/A & N/A & $0.038$ & $0.981$ & $0.946$ & $0.993$ & $0.973$ \\
        & & SplArt & $41.59$ & $0.993$ & $0.0148$ & $0.082$ & $0.990$ & $0.935$ & $0.999$ & $0.974$ \\
        \dashrule{2-11}
        & \multirow{3}{*}{scissor} & PARIS & F & F & F & F & F & F & F & F \\
        & & DTA$^\dagger$ & N/A & N/A & N/A & $0.051$ & $0.863$ & $0.876$ & $0.991$ & $0.910$ \\
        & & SplArt & $46.15$ & $0.996$ & $0.0030$ & $0.044$ & $0.951$ & $0.954$ & $0.999$ & $0.968$ \\
        \dashrule{2-11}
        & \multirow{3}{*}{stapler} & PARIS & F & F & F & F & F & F & F & F \\
        & & DTA$^\dagger$ & N/A & N/A & N/A & $0.035$ & $0.902$ & $0.894$ & $0.994$ & $0.930$ \\
        & & SplArt & $44.81$ & $0.994$ & $0.0021$ & $0.034$ & $0.959$ & $0.940$ & $1.000$ & $0.966$ \\
        \dashrule{2-11}
        & \multirow{3}{*}{washer} & PARIS & F & F & F & F & F & F & F & F \\
        & & DTA$^\dagger$ & N/A & N/A & N/A & $0.016$ & $0.991$ & $0.888$ & $0.998$ & $0.959$ \\
        & & SplArt & $43.75$ & $0.996$ & $0.0079$ & $0.131$ & $0.995$ & $0.930$ & $0.999$ & $0.975$ \\
        \dashrule{2-11}
        & \multirow{3}{*}{mean} & PARIS & $32.21$ & $0.954$ & $0.0854$ & $0.093$ & $0.954$ & $0.914$ & $0.996$ & $0.955$ \\
        & & DTA$^\dagger$ & N/A & N/A & N/A & $0.031$ & $0.926$ & $0.904$ & $0.994$ & $0.941$ \\
        & & SplArt & $43.53$ & $0.995$ & $0.0078$ & $0.053$ & $0.972$ & $0.951$ & $0.999$ & $0.974$ \\
        \cmidrule(lr){1-11}
        \multirow{9}{*}{\rotatebox{90}{Prismatic}} & \multirow{3}{*}{blade} & PARIS & F & F & F & F & F & F & F & F \\
        & & DTA$^\dagger$ & N/A & N/A & N/A & $0.116$ & $0.728$ & $0.416$ & $0.998$ & $0.714$ \\
        & & SplArt & $46.15$ & $0.998$ & $0.0016$ & $0.052$ & $0.900$ & $0.657$ & $1.000$ & $0.852$ \\
        \dashrule{2-11}
        & \multirow{3}{*}{storage} & PARIS & $33.75$ & $0.944$ & $0.1228$ & $0.108$ & $0.954$ & $0.722$ & $0.996$ & $0.891$ \\
        & & DTA$^\dagger$ & N/A & N/A & N/A & $0.017$ & $0.984$ & $0.944$ & $0.997$ & $0.975$ \\
        & & SplArt & $42.62$ & $0.985$ & $0.0481$ & $0.037$ & $0.951$ & $0.850$ & $0.999$ & $0.933$ \\
        \dashrule{2-11}
        & \multirow{3}{*}{mean} & PARIS & $33.75$ & $0.944$ & $0.1228$ & $0.108$ & $0.954$ & $0.722$ & $0.996$ & $0.891$ \\
        & & DTA$^\dagger$ & N/A & N/A & N/A & $0.066$ & $0.856$ & $0.680$ & $0.997$ & $0.844$ \\
        & & SplArt & $44.38$ & $0.991$ & $0.0248$ & $0.044$ & $0.925$ & $0.754$ & $0.999$ & $0.893$ \\
        \cmidrule(lr){1-11}
        \multicolumn{2}{c}{\multirow{3}{*}{Overall}} & PARIS & $32.52$ & $0.952$ & $0.0929$ & $0.096$ & $0.954$ & $0.875$ & $0.996$ & $0.942$ \\
        & & DTA$^\dagger$ & N/A & N/A & N/A & $0.038$ & $0.912$ & $0.859$ & $0.994$ & $0.922$ \\
        & & SplArt & $43.70$ & $0.994$ & $0.0112$ & $0.052$ & $0.963$ & $0.912$ & $0.999$ & $0.958$ \\
        \bottomrule
    \end{tabular}
    \caption{PARIS-PMS Novel View Synthesis Metrics. $^\dagger$DTA requires ground-truth depth.}
    \label{tab:paris-pms-synthesis}
\end{table*}

\begin{table*}[ht]
    \centering
    \begin{tabular}{llcS[table-format=3.2(3.2)]S[table-format=2.2(2.2)]S[table-format=3.2(3.2)]}
        \toprule
        Scene & Method & \makecell[l]{Success \\ Rate} $\big\uparrow$ & {\makecell[l]{$\textrm{err}_a$ \\ ($\times10^{-2}$ DEG)} $\big\downarrow$} & {\makecell[l]{$\textrm{err}_p$ \\ $(\times10^{-3})$} $\big\downarrow$} & {\makecell[l]{$\textrm{err}_r$ \\ ($\times10^{-2}$ DEG)} $\big\downarrow$} \\
        \midrule
        \multirow{3}{*}{\makecell[l]{2230 \\ Chair}} & PARIS & $1/10$ & 44.97 (0.00) & 4.41 (0.00) & 130.66 (0.00) \\
        & DTA$^\dagger$ & $9/10$ & 2.77 (0.79) & 0.18 (0.14) & 5.62 (2.35) \\
        & SplArt & $10/10$ & 0.33 (0.13) & 0.06 (0.03) & 0.84 (0.36) \\
        \dashrule{1-6}
        \multirow{3}{*}{\makecell[l]{5477 \\ Display}} & PARIS & $8/10$ & 86.76 (27.41) & 3.42 (3.21) & 133.12 (43.26) \\
        & DTA$^\dagger$ & $10/10$ & 1.38 (0.39) & 0.27 (0.06) & 2.29 (0.55) \\
        & SplArt & $10/10$ & 1.91 (0.43) & 0.49 (0.25) & 4.10 (0.75) \\
        \dashrule{1-6}
        \multirow{3}{*}{\makecell[l]{7054 \\ Clock}} & PARIS & $0/10$ & F & F & F \\
        & DTA$^\dagger$ & $0/10$ & F & F & F \\
        & SplArt & $1/10$ & 211.22 (0.00) & 1.06 (0.00) & 390.40 (0.00) \\
        \dashrule{1-6}
        \multirow{3}{*}{\makecell[l]{11951 \\ TrashCan}} & PARIS & $0/10$ & F & F & F \\
        & DTA$^\dagger$ & $0/10$ & F & F & F \\
        & SplArt & $10/10$ & 1.65 (0.71) & 21.55 (16.33) & 473.55 (352.80) \\
        \dashrule{1-6}
        \multirow{3}{*}{\makecell[l]{100247 \\ Box}} & PARIS & $0/10$ & F & F & F \\
        & DTA$^\dagger$ & $0/10$ & F & F & F \\
        & SplArt & $8/10$ & 0.79 (0.31) & 0.14 (0.10) & 1.99 (0.67) \\
        \dashrule{1-6}
        \multirow{3}{*}{\makecell[l]{100460 \\ Bucket}} & PARIS & $0/10$ & F & F & F \\
        & DTA$^\dagger$ & $10/10$ & 15.09 (6.89) & 0.61 (0.34) & 24.57 (10.06) \\
        & SplArt & $10/10$ & 0.50 (0.23) & 0.03 (0.04) & 0.79 (0.36) \\
        \dashrule{1-6}
        \multirow{3}{*}{\makecell[l]{100756 \\ Globe}} & PARIS & $0/10$ & F & F & F \\
        & DTA$^\dagger$ & $0/10$ & F & F & F \\
        & SplArt & $1/10$ & 0.72 (0.00) & 0.09 (0.00) & 3.31 (0.00) \\
        \dashrule{1-6}
        \multirow{3}{*}{\makecell[l]{100794 \\ Globe}} & PARIS & $1/10$ & 126.06 (0.00) & 14.17 (0.00) & 533.60 (0.00) \\
        & DTA$^\dagger$ & $10/10$ & 4.59 (3.84) & 2.67 (1.39) & 108.22 (10.35) \\
        & SplArt & $10/10$ & 1.22 (0.35) & 0.12 (0.10) & 2.09 (0.50) \\
        \dashrule{1-6}
        \multirow{3}{*}{\makecell[l]{100882 \\ Switch}} & PARIS & $2/10$ & 393.07 (86.79) & 15.78 (4.03) & 342.44 (171.13) \\
        & DTA$^\dagger$ & $1/10$ & 139.96 (0.00) & 2.67 (0.00) & 63.28 (0.00) \\
        & SplArt & $10/10$ & 19.37 (1.29) & 0.62 (0.23) & 11.20 (0.61) \\
        \dashrule{1-6}
        \multirow{3}{*}{\makecell[l]{101542 \\ Dispenser}} & PARIS & $7/10$ & 71.74 (49.72) & 5.44 (6.19) & 251.86 (267.19) \\
        & DTA$^\dagger$ & $0/10$ & F & F & F \\
        & SplArt & $8/10$ & 0.75 (0.43) & 0.19 (0.06) & 11.48 (0.70) \\
        \dashrule{1-6}
        \multirow{3}{*}{\makecell[l]{102400 \\ Knife}} & PARIS & $0/10$ & F & F & F \\
        & DTA$^\dagger$ & $0/10$ & F & F & F \\
        & SplArt & $5/10$ & 9.01 (5.46) & 22.59 (11.58) & 37.95 (15.46) \\
        \dashrule{1-6}
        \multirow{3}{*}{\makecell[l]{103031 \\ CoffeeMachine}} & PARIS & $3/10$ & 281.43 (90.89) & 12.95 (8.94) & 230.56 (65.56) \\
        & DTA$^\dagger$ & $0/10$ & F & F & F \\
        & SplArt & $10/10$ & 1.69 (0.40) & 0.22 (0.14) & 1.98 (0.58) \\
        \dashrule{1-6}
        \multirow{3}{*}{mean} & PARIS & $18.3\%$ & 167.34 (42.47) & 9.36 (3.73) & 270.37 (91.19) \\
        & DTA$^\dagger$ & $33.3\%$ & 32.76 (2.38) & 1.28 (0.39) & 40.80 (4.66) \\
        & SplArt & $77.5\%$ & 20.76 (0.81) & 3.93 (2.41) & 78.31 (31.06) \\
        \bottomrule
    \end{tabular}
    \caption{\alg-PMS Articulation Metrics on Revolute Scenes. $^\dagger$DTA requires ground-truth depth.}
    \label{tab:splart-pms-articulation-revolute}
\end{table*}

\begin{table*}[ht]
    \centering
    \begin{tabular}{llcS[table-format=3.3(5)]S[table-format=2.3(3)]}
        \toprule
        Scene & Method & \makecell[l]{Success \\ Rate} $\big\uparrow$ & {\makecell[l]{$\textrm{err}_a$ \\ ($\times10^{-2}$ DEG)} $\big\downarrow$} & {\makecell[l]{$\textrm{err}_t$ \\ $(\times10^{-3})$} $\big\downarrow$} \\
        \midrule
        \multirow{3}{*}{\makecell[l]{3558 \\ Bottle}} & PARIS & $5/10$ & 45.83 (22.38) & 23.32 (12.29) \\
        & DTA$^\dagger$ & $10/10$ & 200.52 (10.32) & 4.60 (0.27) \\
        & SplArt & $10/10$ & 3.82 (0.52) & 0.10 (0.02) \\
        \dashrule{1-5}
        \multirow{3}{*}{\makecell[l]{12085 \\ Dishwasher}} & PARIS & $6/10$ & 15.60 (7.00) & 4.67 (1.10) \\
        & DTA$^\dagger$ & $10/10$ & 3.78 (1.96) & 2.46 (0.18) \\
        & SplArt & $10/10$ & 1.00 (0.14) & 0.24 (0.02) \\
        \dashrule{1-5}
        \multirow{3}{*}{\makecell[l]{27189 \\ Table}} & PARIS & $3/10$ & 23.69 (10.43) & 21.18 (3.03) \\
        & DTA$^\dagger$ & $10/10$ & 15.90 (2.43) & 2.69 (0.45) \\
        & SplArt & $10/10$ & 0.14 (0.05) & 0.14 (0.01) \\
        \dashrule{1-5}
        \multirow{3}{*}{\makecell[l]{100248 \\ Suitcase}} & PARIS & $0/10$ & F & F \\
        & DTA$^\dagger$ & $10/10$ & 192.55 (30.46) & 10.57 (1.60) \\
        & SplArt & $10/10$ & 1.22 (0.58) & 0.10 (0.04) \\
        \dashrule{1-5}
        \multirow{3}{*}{\makecell[l]{101713 \\ Pen}} & PARIS & $0/10$ & F & F \\
        & DTA$^\dagger$ & $10/10$ & 179.16 (51.33) & 3.76 (1.04) \\
        & SplArt & $10/10$ & 244.39 (172.86) & 7.96 (2.71) \\
        \dashrule{1-5}
        \multirow{3}{*}{\makecell[l]{102016 \\ USB}} & PARIS & $2/10$ & 17.52 (10.01) & 9.53 (1.87) \\
        & DTA$^\dagger$ & $10/10$ & 44.11 (5.59) & 7.34 (0.69) \\
        & SplArt & $10/10$ & 0.71 (0.33) & 0.49 (0.09) \\
        \dashrule{1-5}
        \multirow{3}{*}{\makecell[l]{102812 \\ Switch}} & PARIS & $8/10$ & 34.54 (18.03) & 9.04 (6.90) \\
        & DTA$^\dagger$ & $10/10$ & 190.87 (8.20) & 38.68 (1.85) \\
        & SplArt & $10/10$ & 1.55 (0.80) & 0.11 (0.05) \\
        \dashrule{1-5}
        \multirow{3}{*}{\makecell[l]{103042 \\ Window}} & PARIS & $5/10$ & 189.87 (54.52) & 30.48 (9.55) \\
        & DTA$^\dagger$ & $10/10$ & 92.93 (1.64) & 11.92 (0.23) \\
        & SplArt & $10/10$ & 1.55 (0.24) & 0.76 (0.18) \\
        \dashrule{1-5}
        \multirow{3}{*}{\makecell[l]{103549 \\ Toaster}} & PARIS & $0/10$ & F & F \\
        & DTA$^\dagger$ & $0/10$ & F & F \\
        & SplArt & $5/10$ & 2.63 (0.74) & 0.30 (0.07) \\
        \dashrule{1-5}
        \multirow{3}{*}{\makecell[l]{103941 \\ Phone}} & PARIS & $10/10$ & 19.39 (10.07) & 6.51 (3.81) \\
        & DTA$^\dagger$ & $10/10$ & 5.06 (0.42) & 1.28 (0.28) \\
        & SplArt & $10/10$ & 0.18 (0.09) & 0.20 (0.03) \\
        \dashrule{1-5}
        \multirow{3}{*}{mean} & PARIS & $39.0\%$ & 49.49 (18.92) & 14.96 (5.51) \\
        & DTA$^\dagger$ & $90.0\%$ & 102.77 (12.48) & 9.25 (0.73) \\
        & SplArt & $95.0\%$ & 25.72 (17.63) & 1.04 (0.32) \\
        \bottomrule
    \end{tabular}
    \caption{\alg-PMS Articulation Metrics on Prismatic Scenes. $^\dagger$DTA requires ground-truth depth.}
    \label{tab:splart-pms-articulation-prismatic}
\end{table*}

\begin{table*}[ht]
    \centering
    \begin{tabular}{llcccccccc}
        \toprule
        Scene & Method & PSNR $\uparrow$ & SSIM $\uparrow$ & LPIPS $\downarrow$ & Depth MAE $\downarrow$ & IoU\textsubscript{s} $\uparrow$ & IoU\textsubscript{m} $\uparrow$ & IoU\textsubscript{bg} $\uparrow$ & mIoU $\uparrow$ \\
        \midrule
        \multirow{3}{*}{\makecell[l]{2230 \\ Chair}} & PARIS & $28.89$ & $0.924$ & $0.1004$ & $0.231$ & $0.510$ & $0.737$ & $0.982$ & $0.743$ \\
        & DTA$^\dagger$ & N/A & N/A & N/A & $0.098$ & $0.764$ & $0.857$ & $0.980$ & $0.867$ \\
        & SplArt & $32.08$ & $0.972$ & $0.0392$ & $0.027$ & $0.498$ & $0.791$ & $0.993$ & $0.761$ \\
        \dashrule{1-10}
        \multirow{3}{*}{\makecell[l]{5477 \\ Display}} & PARIS & $34.34$ & $0.943$ & $0.0979$ & $0.100$ & $0.933$ & $0.973$ & $0.997$ & $0.968$ \\
        & DTA$^\dagger$ & N/A & N/A & N/A & $0.020$ & $0.783$ & $0.968$ & $0.996$ & $0.916$ \\
        & SplArt & $38.33$ & $0.970$ & $0.0379$ & $0.020$ & $0.804$ & $0.971$ & $0.999$ & $0.924$ \\
        \dashrule{1-10}
        \multirow{3}{*}{\makecell[l]{7054 \\ Clock}} & PARIS & F & F & F & F & F & F & F & F \\
        & DTA$^\dagger$ & F & F & F & F & F & F & F & F \\
        & SplArt & $35.54$ & $0.985$ & $0.0187$ & $0.027$ & $0.994$ & $0.811$ & $0.999$ & $0.935$ \\
        \dashrule{1-10}
        \multirow{3}{*}{\makecell[l]{11951 \\ TrashCan}} & PARIS & F & F & F & F & F & F & F & F \\
        & DTA$^\dagger$ & F & F & F & F & F & F & F & F \\
        & SplArt & $30.01$ & $0.952$ & $0.0560$ & $0.083$ & $0.920$ & $0.735$ & $0.980$ & $0.878$ \\
        \dashrule{1-10}
        \multirow{3}{*}{\makecell[l]{100247 \\ Box}} & PARIS & F & F & F & F & F & F & F & F \\
        & DTA$^\dagger$ & F & F & F & F & F & F & F & F \\
        & SplArt & $36.96$ & $0.968$ & $0.0409$ & $0.030$ & $0.989$ & $0.987$ & $0.999$ & $0.992$ \\
        \dashrule{1-10}
        \multirow{3}{*}{\makecell[l]{100460 \\ Bucket}} & PARIS & F & F & F & F & F & F & F & F \\
        & DTA$^\dagger$ & N/A & N/A & N/A & $0.081$ & $0.948$ & $0.455$ & $0.984$ & $0.795$ \\
        & SplArt & $37.83$ & $0.970$ & $0.0397$ & $0.016$ & $0.989$ & $0.832$ & $0.999$ & $0.940$ \\
        \dashrule{1-10}
        \multirow{3}{*}{\makecell[l]{100756 \\ Globe}} & PARIS & F & F & F & F & F & F & F & F \\
        & DTA$^\dagger$ & F & F & F & F & F & F & F & F \\
        & SplArt & $36.30$ & $0.990$ & $0.0346$ & $0.035$ & $0.577$ & $0.346$ & $0.997$ & $0.640$ \\
        \dashrule{1-10}
        \multirow{3}{*}{\makecell[l]{100794 \\ Globe}} & PARIS & $29.45$ & $0.915$ & $0.0965$ & $0.115$ & $0.974$ & $0.937$ & $0.996$ & $0.969$ \\
        & DTA$^\dagger$ & N/A & N/A & N/A & $0.046$ & $0.961$ & $0.943$ & $0.994$ & $0.966$ \\
        & SplArt & $35.47$ & $0.980$ & $0.0216$ & $0.012$ & $0.989$ & $0.985$ & $0.999$ & $0.991$ \\
        \dashrule{1-10}
        \multirow{3}{*}{\makecell[l]{100882 \\ Switch}} & PARIS & $40.23$ & $0.991$ & $0.0418$ & $0.082$ & $0.989$ & $0.822$ & $0.999$ & $0.937$ \\
        & DTA$^\dagger$ & N/A & N/A & N/A & $0.020$ & $0.976$ & $0.676$ & $0.998$ & $0.883$ \\
        & SplArt & $44.04$ & $0.997$ & $0.0174$ & $0.033$ & $0.990$ & $0.835$ & $0.999$ & $0.941$ \\
        \dashrule{1-10}
        \multirow{3}{*}{\makecell[l]{101542 \\ Dispenser}} & PARIS & $31.32$ & $0.966$ & $0.0450$ & $0.105$ & $0.883$ & $0.623$ & $0.998$ & $0.835$ \\
        & DTA$^\dagger$ & F & F & F & F & F & F & F & F \\
        & SplArt & $36.53$ & $0.992$ & $0.0153$ & $0.020$ & $0.869$ & $0.585$ & $0.999$ & $0.818$ \\
        \dashrule{1-10}
        \multirow{3}{*}{\makecell[l]{102400 \\ Knife}} & PARIS & F & F & F & F & F & F & F & F \\
        & DTA$^\dagger$ & F & F & F & F & F & F & F & F \\
        & SplArt & $36.21$ & $0.990$ & $0.0215$ & $0.041$ & $0.986$ & $0.874$ & $0.997$ & $0.952$ \\
        \dashrule{1-10}
        \multirow{3}{*}{\makecell[l]{103031 \\ CoffeeMachine}} & PARIS & $32.29$ & $0.979$ & $0.0659$ & $0.116$ & $0.977$ & $0.825$ & $0.997$ & $0.933$ \\
        & DTA$^\dagger$ & F & F & F & F & F & F & F & F \\
        & SplArt & $37.20$ & $0.993$ & $0.0284$ & $0.034$ & $0.994$ & $0.975$ & $0.999$ & $0.990$ \\
        \dashrule{1-10}
        \multirow{3}{*}{mean} & PARIS & $32.75$ & $0.953$ & $0.0746$ & $0.125$ & $0.878$ & $0.820$ & $0.995$ & $0.897$ \\
        & DTA$^\dagger$ & N/A & N/A & N/A & $0.053$ & $0.886$ & $0.780$ & $0.990$ & $0.886$ \\
        & SplArt & $36.38$ & $0.980$ & $0.0309$ & $0.032$ & $0.883$ & $0.811$ & $0.997$ & $0.897$ \\
        \bottomrule
    \end{tabular}
    \caption{\alg-PMS Novel View and Articulation Synthesis Metrics on Revolute Scenes. $^\dagger$DTA requires ground-truth depth.}
    \label{tab:splart-pms-synthesis-revolute}
\end{table*}

\begin{table*}[ht]
    \centering
    \begin{tabular}{llcccccccc}
        \toprule
        Scene & Method & PSNR $\uparrow$ & SSIM $\uparrow$ & LPIPS $\downarrow$ & Depth MAE $\downarrow$ & IoU\textsubscript{s} $\uparrow$ & IoU\textsubscript{m} $\uparrow$ & IoU\textsubscript{bg} $\uparrow$ & mIoU $\uparrow$ \\
        \midrule
        \multirow{3}{*}{\makecell[l]{3558 \\ Bottle}} & PARIS & $35.75$ & $0.981$ & $0.0276$ & $0.120$ & $0.976$ & $0.676$ & $0.999$ & $0.883$ \\
        & DTA$^\dagger$ & N/A & N/A & N/A & $0.026$ & $0.929$ & $0.580$ & $0.999$ & $0.836$ \\
        & SplArt & $42.05$ & $0.996$ & $0.0076$ & $0.034$ & $0.994$ & $0.968$ & $1.000$ & $0.987$ \\
        \dashrule{1-10}
        \multirow{3}{*}{\makecell[l]{12085 \\ Dishwasher}} & PARIS & $30.81$ & $0.954$ & $0.0927$ & $0.111$ & $0.964$ & $0.929$ & $0.994$ & $0.962$ \\
        & DTA$^\dagger$ & N/A & N/A & N/A & $0.113$ & $0.873$ & $0.773$ & $0.982$ & $0.876$ \\
        & SplArt & $35.70$ & $0.988$ & $0.0388$ & $0.019$ & $0.985$ & $0.965$ & $0.998$ & $0.983$ \\
        \dashrule{1-10}
        \multirow{3}{*}{\makecell[l]{27189 \\ Table}} & PARIS & $29.55$ & $0.887$ & $0.1618$ & $0.131$ & $0.958$ & $0.866$ & $0.995$ & $0.940$ \\
        & DTA$^\dagger$ & N/A & N/A & N/A & $0.044$ & $0.964$ & $0.889$ & $0.992$ & $0.948$ \\
        & SplArt & $34.76$ & $0.963$ & $0.0560$ & $0.035$ & $0.979$ & $0.949$ & $0.998$ & $0.975$ \\
        \dashrule{1-10}
        \multirow{3}{*}{\makecell[l]{100248 \\ Suitcase}} & PARIS & F & F & F & F & F & F & F & F \\
        & DTA$^\dagger$ & N/A & N/A & N/A & $0.025$ & $0.986$ & $0.574$ & $0.998$ & $0.853$ \\
        & SplArt & $39.76$ & $0.995$ & $0.0104$ & $0.038$ & $0.995$ & $0.934$ & $0.999$ & $0.976$ \\
        \dashrule{1-10}
        \multirow{3}{*}{\makecell[l]{101713 \\ Pen}} & PARIS & F & F & F & F & F & F & F & F \\
        & DTA$^\dagger$ & N/A & N/A & N/A & $0.037$ & $0.956$ & $0.319$ & $0.999$ & $0.758$ \\
        & SplArt & $40.85$ & $0.998$ & $0.0135$ & $0.028$ & $0.981$ & $0.789$ & $1.000$ & $0.923$ \\
        \dashrule{1-10}
        \multirow{3}{*}{\makecell[l]{102016 \\ USB}} & PARIS & $31.45$ & $0.970$ & $0.1005$ & $0.191$ & $0.731$ & $0.583$ & $0.994$ & $0.769$ \\
        & DTA$^\dagger$ & N/A & N/A & N/A & $0.157$ & $0.710$ & $0.553$ & $0.979$ & $0.747$ \\
        & SplArt & $35.08$ & $0.987$ & $0.0475$ & $0.054$ & $0.659$ & $0.533$ & $0.998$ & $0.730$ \\
        \dashrule{1-10}
        \multirow{3}{*}{\makecell[l]{102812 \\ Switch}} & PARIS & $35.23$ & $0.984$ & $0.0433$ & $0.102$ & $0.987$ & $0.908$ & $0.998$ & $0.964$ \\
        & DTA$^\dagger$ & N/A & N/A & N/A & $0.024$ & $0.970$ & $0.616$ & $0.996$ & $0.861$ \\
        & SplArt & $38.69$ & $0.995$ & $0.0235$ & $0.031$ & $0.995$ & $0.971$ & $0.999$ & $0.988$ \\
        \dashrule{1-10}
        \multirow{3}{*}{\makecell[l]{103042 \\ Window}} & PARIS & $29.11$ & $0.967$ & $0.0808$ & $0.127$ & $0.859$ & $0.631$ & $0.996$ & $0.829$ \\
        & DTA$^\dagger$ & N/A & N/A & N/A & $0.088$ & $0.900$ & $0.744$ & $0.991$ & $0.878$ \\
        & SplArt & $32.35$ & $0.984$ & $0.0499$ & $0.039$ & $0.868$ & $0.695$ & $0.999$ & $0.854$ \\
        \dashrule{1-10}
        \multirow{3}{*}{\makecell[l]{103549 \\ Toaster}} & PARIS & F & F & F & F & F & F & F & F \\
        & DTA$^\dagger$ & F & F & F & F & F & F & F & F \\
        & SplArt & $39.04$ & $0.994$ & $0.0195$ & $0.036$ & $0.997$ & $0.949$ & $0.999$ & $0.982$ \\
        \dashrule{1-10}
        \multirow{3}{*}{\makecell[l]{103941 \\ Phone}} & PARIS & $34.21$ & $0.972$ & $0.0592$ & $0.089$ & $0.938$ & $0.956$ & $0.998$ & $0.964$ \\
        & DTA$^\dagger$ & N/A & N/A & N/A & $0.027$ & $0.913$ & $0.952$ & $0.997$ & $0.954$ \\
        & SplArt & $39.42$ & $0.993$ & $0.0219$ & $0.020$ & $0.971$ & $0.980$ & $0.999$ & $0.984$ \\
        \dashrule{1-10}
        \multirow{3}{*}{mean} & PARIS & $32.30$ & $0.959$ & $0.0808$ & $0.124$ & $0.916$ & $0.793$ & $0.996$ & $0.902$ \\
        & DTA$^\dagger$ & N/A & N/A & N/A & $0.060$ & $0.911$ & $0.667$ & $0.993$ & $0.857$ \\
        & SplArt & $37.77$ & $0.989$ & $0.0288$ & $0.033$ & $0.942$ & $0.873$ & $0.999$ & $0.938$ \\
        \bottomrule
    \end{tabular}
    \caption{\alg-PMS Novel View and Articulation Synthesis Metrics on Prismatic Scenes. $^\dagger$DTA requires ground-truth depth.}
    \label{tab:splart-pms-synthesis-prismatic}
\end{table*}

\end{document}